\documentclass{article}

    \usepackage[preprint]{neurips_2025}

\usepackage[utf8]{inputenc} 
\usepackage[T1]{fontenc}    
\usepackage{hyperref}       
\usepackage{url}            
\usepackage{booktabs}       
\usepackage{amsfonts}       
\usepackage{nicefrac}       
\usepackage{microtype}      
\usepackage{xcolor}         

\usepackage{amsmath}
\usepackage{amssymb}
\usepackage{mathtools}
\usepackage{amsthm}

\usepackage[capitalize,noabbrev]{cleveref}

\theoremstyle{plain}

\theoremstyle{definition}

\theoremstyle{remark}

\usepackage[textsize=tiny]{todonotes}

\usepackage[subfigure]{tocloft}
\newcommand{\listappendixname}{}
\newlistof{appendix}{apc}{\listappendixname}

\usepackage{subfigure}
\usepackage{wrapfig}
\usepackage{graphicx}
\usepackage{subfigure}
\usepackage{multirow}
\usepackage{algorithm}
\usepackage{algorithmic}

\title{Leave it to the Specialist: Repair Sparse LLMs with Sparse Fine-Tuning via Sparsity Evolution}

\author{ Qiao Xiao$^{1}$, Alan Ansell$^{2}$, Boqian Wu$^{3, 4}$, Lu Yin$^{5}$, Mykola Pechenizkiy$^{1}$, \and \textbf{Shiwei Liu$^{6, 7, 8}$, Decebal Constantin Mocanu$^{3}$}  \\
$^1$Eindhoven University of Technology, $^2$University of Cambridge, \\ $^3$University of Luxembourg, $^4$University of Twente,  $^5$University of Surrey, \\ $^6$Tübingen AI Center, $^7$Max Planck Institute for Intelligent Systems, $^8$ELLIS Institute Tübingen \\
Correspondence to: \texttt{q.xiao@tue.nl, b.wu@utwente.nl}
}

\begin{document}

\maketitle


\begin{abstract}
Sparse large language models (LLMs) offer an attractive direction toward efficient deployment, but adapting them to downstream tasks remains challenging. The central difficulty is to enable effective task adaptation without sacrificing the efficiency advantages of sparsity. Existing fine-tuning methods are not well-suited to this setting, as they either introduce additional dense parameters or assume a fixed sparse topology, limiting their compatibility with sparse LLMs.
In this paper, we propose \textbf{Sparsity Evolution Fine-Tuning (SEFT)}, a fine-tuning framework designed specifically for sparse LLMs. SEFT allows sparse structure to evolve during fine-tuning by periodically reallocating sparse task-specific updates and reactivating previously pruned weights when beneficial. At the same time, SEFT preserves the efficiency advantages of sparsity through topology adaptation based on parameter importance.
Experiments on LLaMA, DeepSeek, and Mistral models across multiple benchmarks show that SEFT delivers stronger performance while offering superior memory and time efficiency compared to existing baselines. Our code is publicly available at: \url{https://github.com/QiaoXiao7282/SEFT}.

\end{abstract}

\section{Introduction}
\label{introduction}

Large language models (LLMs) have achieved remarkable success across a wide range of tasks, including language understanding \citep{zhu2023minigpt}, reasoning \citep{wei2022chain}, and code generation \citep{vaithilingam2022expectation}. However, their massive parameter counts create substantial challenges for real-world deployment, especially in memory-constrained and latency-sensitive settings. As a result, improving the efficiency of LLMs has become a central problem in both research and practice.

Among existing efficiency techniques, post-training pruning offers an attractive path for compressing LLMs without the prohibitive cost of retraining from scratch \citep{ma2023llm, zhu2024survey}. Recent methods such as SparseGPT \citep{frantar2023sparsegpt} and Wanda \citep{sun2023simple} can produce highly sparse models while preserving much of the original performance. Yet pruning alone is not sufficient: as sparsity increases, performance degradation becomes increasingly pronounced, especially in high-sparsity regimes (e.g., $\geq 60\%$~\citep{lu2024alphapruning}). More importantly, post-training pruning typically produces a \emph{task-agnostic} sparse structure, determined before downstream adaptation. A sparse topology that appears effective under calibration data may still remove connections that are important for a specific downstream task~\citep{bandari2024c4, williams2026compressing}.

\begin{figure}[!t]
    \centering
    \vskip -0.2in
    \includegraphics[width=0.82\textwidth]{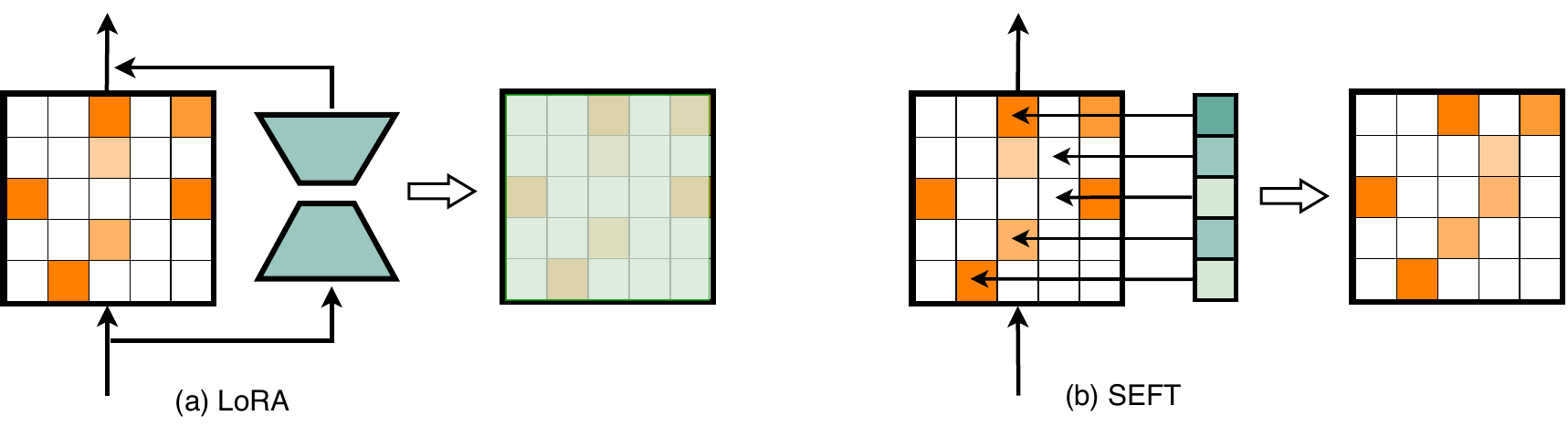}
    \vskip -0.1in
    \caption{Comparison of LoRA and SEFT fine-tuning for sparse LLMs. While LoRA-based methods require low-rank matrices, SEFT utilizes indices (arrows) and corresponding deltas (green squares) to update LLM parameters. 
    }
    \vskip -0.1in
    \label{fig: sft_intro}
\end{figure}

This exposes a fundamental limitation in current adaptation methods on sparse LLMs: the challenge is not only to preserve the efficiency benefits of sparsity, but also to adapt sparse models to downstream tasks when the pruned topology itself may be misaligned with task-specific needs. However, existing methods do not fully address this challenge. General parameter-efficient fine-tuning (PEFT) methods, such as LoRA and its variants \citep{hu2021lora, xu2023qa, dettmers2024qlora}, are effective for adapting dense LLMs, but are not well suited to sparse deployment. As they either introduce additional storage overhead for adapters or produce dense models after merging, thereby undermining the efficiency benefits of sparsity, as illustrated in Figure~\ref{fig: sft_intro}(a). More recent sparsity-preserved fine-tuning methods \citep{luspp, munoz2024sqft, hu2025lors} maintain sparsity throughout adaptation, but typically update only a predefined set of sparse parameters. This restriction can lead to suboptimal performance when the resulting sparse topology is not well aligned with the downstream task.

As a result, current methods either fail to preserve sparse efficiency or do not support the sparse topology itself to adapt. This raises a key question: \emph{can we efficiently fine-tune sparse LLMs while allowing their sparse updates to evolve for better adaptation to downstream tasks?}

To answer this question, we propose \textbf{Sparsity Evolution Fine-Tuning (SEFT)}, a sparse fine-tuning framework for adapting post-pruning sparse LLMs. The core idea of SEFT is to allow sparse task-specific updates evolve with downstream tasks in both their locations and mask constraints. First, the locations of sparse task-specific updates are periodically refreshed during fine-tuning. Second, these update locations are not restricted to the initial pruning mask, allowing previously pruned weights to be explored and reactivated when beneficial. Together, these properties enable SEFT to better align sparse LLMs with downstream tasks, as illustrated in Figure~\ref{fig: sft_intro}(b).

To realize this efficiently, SEFT consists of two coupled components. A lightweight \textit{Delta Support Update} governs the evolution of sparse task-specific updates by refreshing where they are allocated, while \textit{Sparse Topology Adaptation} preserves the efficiency advantages of sparsity by enforcing the target sparsity level on the model. Unlike LoRA-based approaches that introduce auxiliary adapters, SEFT directly updates the model while keeping the base model sparse.

We evaluate SEFT on a diverse set of sparse LLMs from the LLaMA \citep{touvron2023llama1, touvron2023llama2, dubey2024llama3}, DeepSeek \citep{bi2024deepseek}, and Mistral \citep{jiang2023mistral7b} families. Across model scales from 7B to 13B and a range of downstream benchmarks, including commonsense reasoning, MMLU, and GSM8K, SEFT consistently delivers better performance while also offering favorable memory usage and training efficiency.

Our contributions in this work can be summarized as follows:

\begin{itemize}
\item We propose SEFT, an efficient fine-tuning approach tailored to sparse LLMs, which dynamically evolves the sparse topology to enable task-specific adaptation.
\end{itemize}

\begin{itemize}
\item We introduce a sparsity adaptation mechanism based on a sensitivity-guided pruning criterion, which preserves the target sparsity level without sacrificing performance.
\end{itemize}

\begin{itemize}
\item We demonstrate that SEFT offers favorable memory usage and training efficiency compared with prior sparsity-preserving fine-tuning methods.
\end{itemize}

\begin{itemize}
\item We conduct comprehensive experiments across multiple sparse LLM architectures (LLaMA, DeepSeek, and Mistral) and tasks (commonsense reasoning, MMLU, and GSM8K), validating SEFT’s generality and effectiveness in both general-purpose and task-specific settings.
\end{itemize}

\section{Preliminaries}
\label{background}

\subsection{Parameter-Efficient Fine-Tuning (PEFT)}
Parameter-efficient fine-tuning (PEFT) has emerged as a popular approach to adapt LLMs to downstream tasks while significantly reducing the computational and memory overhead associated with traditional dense full fine-tuning \citep{pfeiffer2023modular, ding2023parameter, han2024parameter}. Instead of updating all parameters of a pre-trained model, PEFT methods introduce additional trainable parameters, which typically are a small fraction of the original model, while keeping the base model weights frozen. This significantly reduces computational overhead during fine-tuning. 

One widely used PEFT method is Low-Rank Adaptation (LoRA) \citep{hu2021lora}. In LoRA, low-rank matrices are introduced into specific layers (e.g., attention or feedforward layers) to capture task-specific adaptations. Specifically, let $\mathbf{W} \in \mathbb{R}^{d \times d}$ represent a frozen weight matrix in a pre-trained model. LoRA introduces a low-rank decomposition:

\begin{equation}
\mathbf{W}^{\prime}=\mathbf{W}+\Delta \mathbf{W}, \quad \Delta \mathbf{W}=\mathbf{A B}
\end{equation}

where $\mathbf{A} \in \mathbb{R}^{d \times r}$ and $\mathbf{B} \in \mathbb{R}^{r \times d}$ are low-rank matrices $(r \ll d)$, and only $\mathbf{A}$ and $\mathbf{B}$ are trainable. 

While PEFT methods like LoRA have proven effective for dense models, they are inherently not well-suited for sparse LLMs, as they either require dense updates or fail to preserve sparsity after fine-tuning.

\subsection{Sparse Fine-Tuning (SFT)}

Unlike LoRA-based approaches, Sparse Fine-Tuning (SFT) \citep{guo-etal-2021-parameter,xu-etal-2021-raise,ansell-etal-2022-composable} directly modifies the base model by introducing task-specific updates in the form of a sparse ``delta” vector $\boldsymbol{\delta} \in \mathbb{R}^{d_\theta}$, which is added to the pre-trained parameters $\boldsymbol{\theta} \in \mathbb{R}^{d_\theta}$. The resulting fine-tuned model can be expressed as:

\begin{equation}
f^{\prime}(\cdot; \boldsymbol{\theta}) = f(\cdot; \boldsymbol{\theta} + \boldsymbol{\delta}),
\end{equation}

Here, $\boldsymbol{\delta}$ contains non-zero updates only at specific locations in $\boldsymbol{\theta}$, allowing for a sparse parameterization that enables efficient fine-tuning in practice.

\textbf{Sparse Parameterization.}
A sparse delta vector $\boldsymbol{\delta} \in \mathbb{R}^{d_\theta}$ with $d_\phi$ non-zero entries can be represented by its support indices and corresponding update values:
\begin{itemize}
    \item \textbf{Support indices} ($\boldsymbol{\eta}$): a vector of unique indices $\boldsymbol{\eta} \in \{1,2,\ldots,d_\theta\}^{d_\phi}$ specifying the parameter locations where sparse updates are applied;
    \item \textbf{Update values} ($\boldsymbol{\phi}$): the corresponding update values at those locations, with $\boldsymbol{\phi} \in \mathbb{R}^{d_\phi}$.
\end{itemize}

Together, $(\boldsymbol{\eta}, \boldsymbol{\phi})$ define the sparse update $\boldsymbol{\delta}$. Here, $d_\phi$ is a user-specified support budget that controls how many parameters are updated during fine-tuning. Since $\boldsymbol{\delta}$ modifies only a small subset of the full parameter vector $\boldsymbol{\theta}$, we have $d_\phi \ll d_\theta$, which makes sparse fine-tuning computationally efficient.

\textbf{Optimization Objective.}
Sparse fine-tuning can be formulated as an optimization problem that jointly determines \emph{where} to apply sparse updates and \emph{what} update values to assign for a given task dataset $\mathcal{D}$. Formally, the objective is

\begin{equation}
\boldsymbol{\eta}^{\star}, \boldsymbol{\phi}^{\star}
=
\underset{\boldsymbol{\eta},\,\boldsymbol{\phi}}{\arg\min}
\ \mathcal{L}(\mathcal{D}; \boldsymbol{\theta}, \boldsymbol{\eta}, \boldsymbol{\phi}),
\end{equation}
where $\mathcal{L}$ is the task loss. This formulation jointly optimizes the support of the sparse delta (through $\boldsymbol{\eta}$) and the corresponding update values (through $\boldsymbol{\phi}$), while keeping the number of updated parameters bounded by the support budget $d_\phi$. Recent work has shown that this sparse optimization framework can scale effectively in dense LLMs \citep{ansell2024scaling}, motivating its extension to sparse LLM adaptation.


\begin{figure*}[!t]
    \centering
    \includegraphics[width=\textwidth]{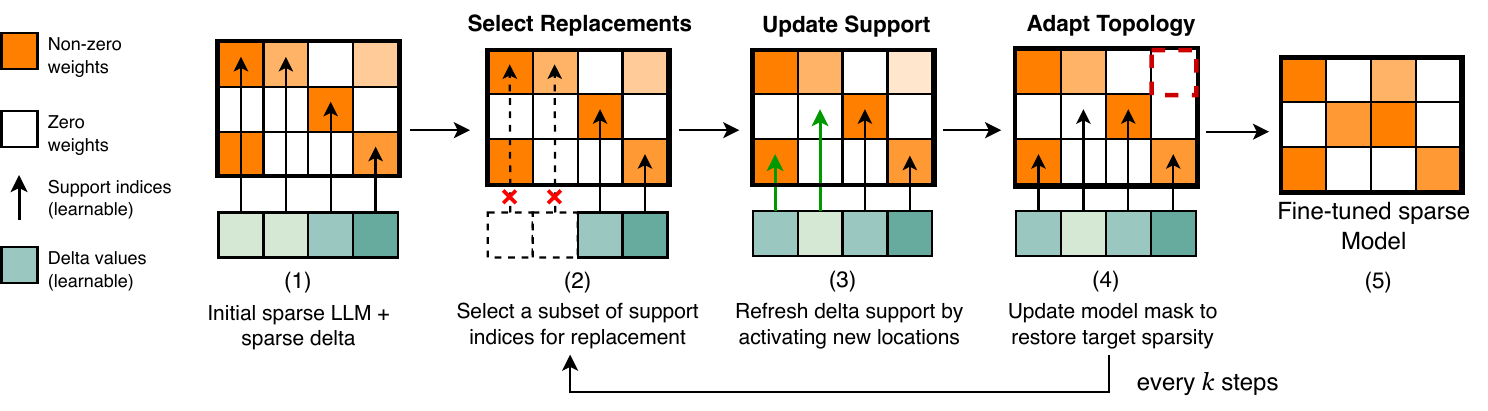}
    \vskip -0.05in
    \caption{An illustration of SEFT for sparse LLMs. The SEFT process consists of: (1) starting with a sparse LLM, along with learnable indices (arrows) and corresponding deltas (green squares) applied to the LLM parameters; (2) dropping obsolete indices (dashed arrows); (3) growing new indices (green arrows); (4) adapting parameters to the target sparsity level (red squares). These steps are repeated every $k$ steps. The final output is a sparse LLM after fine-tuning.}
    \vskip -0.05in
    \label{fig: sft_method}
\end{figure*}

\section{Sparsity Evolution Fine-Tuning for Sparse LLMs}
\label{method}

Building on the sparse fine-tuning paradigm, we propose Sparsity Evolution Fine-Tuning (SEFT), a method for adapting post-pruning sparse LLMs by allowing sparse topology to evolve during task-specific fine-tuning. Unlike conventional sparse fine-tuning, which is primarily developed for dense pre-trained models, SEFT is tailored to sparse models and dynamically evolves updates in both locations and the set of active model weights. This enables effective downstream adaptation while maintaining sparsity efficiency, as illustrated in Figure~\ref{fig: sft_method}.

Let $\boldsymbol{\theta} \in \mathbb{R}^{d_\theta}$ denote the pre-trained dense model weights. After post-training pruning, the resulting sparse model is
$\boldsymbol{\theta}_0^{\prime}=\boldsymbol{\theta} \odot \boldsymbol{M}_0$,
where $\boldsymbol{M}_0 \in\{0,1\}^{d_\theta}$ is the initial binary mask, and $M_0^i=1$ and $M_0^i=0$ indicate active and pruned parameters, respectively. The target sparsity level $\rho$ is defined as the fraction of inactive parameters.
SEFT fine-tunes the sparse LLM model through a learnable sparse delta $\boldsymbol{\delta}_t$:

\begin{equation}
f^{\prime}\left(\cdot ; \boldsymbol{\theta}_t^{\prime}\right)=f\left(\cdot ; \boldsymbol{\theta}_t^{\prime}+\boldsymbol{\delta}_t\right),
\end{equation}
where $\boldsymbol{\theta}_t^{\prime}=\boldsymbol{\theta} \odot \boldsymbol{M}_t$ denotes the sparse model weights at step $t$. Notably, SEFT does not restrict $\boldsymbol{\delta}_t$ to the support of the initial mask $\boldsymbol{M}_0$, allowing sparse updates to move across parameter locations, including previously pruned weights.

To realize this sparsity evolution, SEFT consists of two complementary components: \textit{Delta Support Update}, which evolves where sparse task-specific updates are allocated, and \textit{Sparse Topology Adaptation}, which updates the sparse model structure while maintaining the target sparsity level $\rho$. Together, these components enable SEFT to directly obtain a fine-tuned sparse model, rather than relying on a separate fine-tune-then-prune pipeline.

\subsection{Delta Support Update}
\label{sec:evolution}

Inspired by Dynamic Sparse Training (DST) \citep{mocanu2018scalable, evci2020rigging}, SEFT periodically updates the support of the sparse delta during fine-tuning. Following the sparse fine-tuning formulation, we represent the sparse update $\boldsymbol{\delta}_t$ at step $t$ by its support indices $\boldsymbol{\eta}_t$ and corresponding update values $\boldsymbol{\phi}_t$, where $\boldsymbol{\eta}_t \in \{1,\ldots,d_\theta\}^{d_\phi}$ specifies the updated parameter locations and $\boldsymbol{\phi}_t \in \mathbb{R}^{d_\phi}$ stores the associated update values.

Every $k$ training steps, SEFT refreshes the support $\boldsymbol{\eta}_t$ while keeping the budget $d_\phi$ fixed:

\begin{equation}
\boldsymbol{\eta}_{t+1}
=
(\boldsymbol{\eta}_t \setminus \boldsymbol{\eta}_{\mathrm{replace}})
\cup
\boldsymbol{\eta}_{\mathrm{new}},
\label{eq:update}
\end{equation}
where $\boldsymbol{\eta}_{\mathrm{replace}} \subset \boldsymbol{\eta}_t$ contains the $\tau(t)$ support indices whose corresponding values in $\boldsymbol{\phi}_t$ have the smallest magnitudes, and $\boldsymbol{\eta}_{\mathrm{new}}$ contains the $\tau(t)$ inactive locations with the largest gradient magnitudes under the current task loss. Here, $\tau(t)$ controls how many support locations are refreshed at step $t$, while the support size remains fixed, i.e.,
 $|\boldsymbol{\eta}_{t+1}|=|\boldsymbol{\eta}_t|=d_\phi$.

This update rule reflects a simple but effective principle: delta locations with small update magnitudes are less useful for the current task, whereas inactive locations with large gradients are promising candidates for future adaptation. By periodically refreshing $\boldsymbol{\eta}_t$, SEFT reallocates its sparse adaptation budget toward more task-relevant parameter locations throughout fine-tuning.

Importantly, this mechanism allows $\boldsymbol{\delta}_t$ to move beyond the initial pruning mask $\boldsymbol{M}_0$ and place updates on both originally retained and previously pruned weights. As a result, the effective sparse structure induced during fine-tuning may temporarily become denser than the initial pruned model, which motivates the subsequent Sparse Topology Adaptation step to restore the target sparsity level.

\subsection{Sparsity Topology Adaptation}
\label{sec:adaptation}

\begin{wrapfigure}{r}{0.55\textwidth}
\vskip -0.4in
\begin{minipage}{0.55\textwidth}
\begin{algorithm}[H]
\caption{Sparsity Evolution Fine-Tuning (SEFT)}
\label{alg:SEFT}
{\small
\begin{algorithmic}
\STATE {\bfseries Input:} Pre-trained weights $\boldsymbol{\theta}$, initial mask $\boldsymbol{M}_0$, target sparsity level $\rho$, sparse delta budget $d_\phi$, total training steps $T$, update frequency $k$.
\STATE {\bfseries Initialize:} $\boldsymbol{M}_1 \leftarrow \boldsymbol{M}_0$, $\boldsymbol{\theta}_1' \leftarrow \boldsymbol{\theta}\odot\boldsymbol{M}_1$, $\boldsymbol{\delta}_1 \leftarrow \mathbf{0}$.
\STATE {\bfseries Output:} Fine-tuned sparse model $\boldsymbol{\theta}_T' + \boldsymbol{\delta}_T$.

\FOR {$t = 1$ to $T$}
    \STATE Update $\boldsymbol{\delta}_t$ using task-specific loss $\mathcal{L}$ on $\boldsymbol{\theta}_t' + \boldsymbol{\delta}_t$.
    
    \IF {$t \bmod k = 0$}
        \STATE \textcolor{blue}{\textbf{Delta Support Update}}
        \STATE \quad Select support indices to be replaced using $\boldsymbol{\phi}_t$.
        \STATE \quad Update support $\boldsymbol{\eta}_t$ by Eq.~\ref{eq:update}.

        \STATE \textcolor{blue}{\textbf{Sparse Topology Adaptation}}
        \STATE \quad Compute importance scores $s_i$ by Eq.~\ref{eq:sensitivity}.
        \STATE \quad Update mask $\boldsymbol{M}_t$ according to Eq.~\ref{eq:mask}.
        \STATE \quad Update sparse model weights: $\boldsymbol{\theta}_t' \leftarrow \boldsymbol{\theta}\odot\boldsymbol{M}_t$.
    \ENDIF
\ENDFOR

\STATE \textbf{Return:} $\boldsymbol{\theta}_T' + \boldsymbol{\delta}_T$.
\end{algorithmic}}
\end{algorithm}
\end{minipage}
\vskip -0.1in
\end{wrapfigure}


While Delta Support Update improves task adaptation by refreshing where sparse updates are applied, it may also place updates on weights that were initially inactive. As a result, the effective model structure during fine-tuning can become temporarily denser than the initial pruned model. To restore the target sparsity level, SEFT further performs \textit{Sparse Topology Adaptation}, which determines which model weights remain active based on parameter importance.

Specifically, SEFT assigns each model parameter an importance score $s_i$, which by default is computed using a sensitivity-based criterion:
\begin{equation}
s_i = \left| \theta_i \nabla_{\theta_i}\mathcal{L} \right|,
\label{eq:sensitivity}
\end{equation}
where $\nabla_{\theta_i}\mathcal{L}$ denotes the gradient of the task loss with respect to parameter $\theta_i$. This score captures both the scale of a parameter and its local influence on the loss. As shown in Appendix~\ref{sec:adapt} and Figure~\ref{fig: ablation}(a), this criterion performs well in our setting and is used throughout the paper unless otherwise specified.

Given the importance scores $\{s_i\}_{i=1}^{d_\theta}$, SEFT updates the model mask by retaining the top $(1-\rho)$ fraction of parameters:
\begin{equation}
M_t^i=
\begin{cases}
1, & \text{if } s_i \text{ is among the largest } (1-\rho)\text{ fraction of all scores},\\
0, & \text{otherwise}.
\end{cases}
\label{eq:mask}
\end{equation}
The sparse model is then updated as
$
\boldsymbol{\theta}_t^{\prime}=\boldsymbol{\theta}\odot\boldsymbol{M}_t.
$

In contrast to Delta Support Update, which operates on the sparse delta $\boldsymbol{\delta}_t$ under a fixed support budget $d_\phi$, Sparse Topology Adaptation operates on the base model $\boldsymbol{\theta}$ and determines which weights remain active under the target sparsity level $\rho$. Together, these two components enable SEFT to adapt sparse LLMs flexibly while preserving the efficiency advantages of sparsity (summarized in Alg.~\ref{alg:SEFT}). In practice, we use accumulated rather than instantaneous gradients to improve training stability.

\section{Experiments and Results}
\label{experiments}

\textbf{Sparse LLMs.} We evaluate on several widely-used open-source LLMs, including the LLaMA family, DeepSeek, and Mistral. These models are pruned using state-of-the-art post-training pruning methods, such as SparseGPT and Wanda, to achieve the desired sparsity levels. Our focus is primarily on unstructured sparsity as it allows for fine-grained control over the parameters and has shown strong empirical results in prior works \citep{frantar2023sparsegpt, sun2023simple, yinoutlier}. We target highly sparse models (e.g., sparsity $\geq$ 0.6), where performance degradation becomes more pronounced, necessitating fine-tuning. Moreover, high sparsity levels are generally more hardware-friendly for practical speedup \citep{gale2020sparse}, as demonstrated by our analysis in Section~\ref{main_speedup}.

\textbf{Fine-tuning Tasks.} To comprehensively evaluate the effectiveness of SEFT, we conduct experiments across two categories of fine-tuning tasks:

(1) General-domain fine-tuning:
This setting assesses the ability of fine-tuning methods to restore model capability after pruning using general-domain data. Specifically, models are fine-tuned on a general pretraining dataset, using 30k randomly selected samples from the C4 corpus \citep{raffel2020exploring}, and evaluated using Wikitext perplexity (PPL) \citep{merity2016pointer} and Commonsense Reasoning benchmark \citep{gao2021framework}.

(2) Supervised fine-tuning:
This setting evaluates the models’ ability to learn task-specific information through supervised data:
(i) Commonsense Reasoning: Models are fine-tuned on a 170k-sample commonsense reasoning dataset and evaluated on zero-shot performance across seven commonsense reasoning tasks: BoolQ, RTE, HellaSwag, WinoGrande, ARC-e, ARC-c, and OBQA, following the evaluation protocol in \citep{hu2023llm}.
(ii) MMLU: The Massive Multitask Language Understanding benchmark \citep{hendrycks2020measuring} includes 57 tasks spanning diverse domains, such as elementary mathematics, US history, computer science, and law. We adopt a zero-shot evaluation protocol, where models are fine-tuned on an auxiliary training dataset and evaluated on MMLU.
(iii) GSM8K \citep{cobbe2021training}: This benchmark consists of grade-school-level math word problems. Models are fine-tuned on the training split of GSM8K and evaluated in a zero-shot setting.


\textbf{Baselines.} To evaluate the effectiveness of SEFT, we first obtain sparse LLMs and subsequently fine-tune them using the following sparsity-preserving baselines:

(1) LoRA*: After LoRA fine-tuning, we apply Wanda pruning post-hoc to restore the desired sparsity level and retain the efficiency.

(2) SPP: This method fine-tunes sparse models using designed sparsity-preserving adapter, allowing the final model to maintain the target sparsity level after merging \citep{luspp}.

(3) SQFT: A sparsity-preserving version from \citep{munoz2024sqft} that excludes quantization of the base model. It maintains the target sparsity during fine-tuning by applying a binary mask.

For fair comparison, all fine-tuned LLMs are restored to target sparsity level prior to evaluation, and all fine-tuning methods are configured with the same number of trainable parameters, matched to that of LoRA under the corresponding rank configuration. We adopt a fixed rank of 32
as the default setting. Additional experimental details are provided in Appendix~\ref{sec:hyper}.

\subsection{Overall Performance}

In this section, we present a comprehensive evaluation of SEFT across a range of fine-tuning scenarios. We begin by analyzing its effectiveness under general-domain, followed by results on supervised fine-tuning tasks. Finally, we assess its efficiency in terms of memory cost, training speed, and inference latency.

\begin{wraptable}{r}{0.55\textwidth}
\vskip -0.4in
\begin{minipage}{0.55\textwidth}
\caption{Performance comparison of different methods with and without fine-tuning applied to sparse LLaMA models at a sparsity level of $\rho=0.7$.}
\resizebox{0.95\linewidth}{!}{
\begin{tabular}{clccc}
\toprule
\textbf{LLaMA} & \textbf{Method}  & \textbf{PPL} ($\downarrow$) & \textbf{LM-eval} ($\uparrow$)  \\
\midrule
\multirow{8}{*}{V2-7B} 
& Wanda            & 75.19 & 34.83 \\
& Wanda+LoRA$^*$   & 11.71 & 44.29  \\
& Wanda+SPP         & 11.53 & 44.46 \\
& Wanda+SQFT        & 11.94 & 44.73 \\
& Wanda+SEFT        & \textbf{11.19} & \textbf{45.61}  \\
\cmidrule{2-4}
& SparseGPT            & 27.31 & 41.51 \\
& SparseGPT+LoRA$^*$  & 12.86 & 45.42  \\
& SparseGPT+SPP        & 11.58 & 47.27  \\
& SparseGPT+SQFT       & 11.71 & 47.09\\
& SparseGPT+SEFT        & \textbf{11.00} & \textbf{47.95}   \\
\midrule
\multirow{8}{*}{V3-8B} 
& Wanda            & 120.20 & 34.92 \\
& Wanda+LoRA$^*$  & 18.28  & 43.32 \\
& Wanda+SPP        & 16.73 & 43.68  \\
& Wanda+SQFT       & 17.16 & 43.56\\
& Wanda+SEFT        & \textbf{16.17} & \textbf{44.55}  \\
\cmidrule{2-4}
& SparseGPT         &43.25 & 41.70 \\
& SparseGPT+LoRA$^*$  & 17.81 & 46.11  \\
& SparseGPT+SPP        & 15.25 & 48.33 \\
& SparseGPT+SQFT       & 16.14 & 47.70\\
& SparseGPT+SEFT        & \textbf{15.09} & \textbf{48.89}  \\
\bottomrule
\end{tabular}
}
\label{tab:recovery}
\end{minipage}
\vskip -0.2in
\end{wraptable}

\subsubsection{General-domain fine-tuning} 

We evaluate the effectiveness of SEFT under general-domain fine-tuning by adapting sparse LLMs on a pretraining dataset, using 30k randomly sampled examples from C4~\citep{raffel2020exploring}. To obtain sparse models, we first apply post-training pruning methods including Wanda and SparseGPT to reach a 70\% sparsity level.

We use two evaluation metrics: LM-eval, which reports zero-shot accuracy across seven tasks from the EleutherAI LM Harness \citep{gao2021framework}, and PPL, which denotes the perplexity on Wikitext-2 \citep{merity2016pointer}. As shown in Table~\ref{tab:recovery}, fine-tuning significantly improves the performance of sparse LLMs, as indicated by lower perplexity and higher LM-eval scores compared to their unfinetuned counterparts.
Notably, SEFT consistently performs strongly across different settings, achieving competitive or better results than existing baselines, particularly LoRA* and SQFT, across both evaluation metrics. These gains are observed across different model scales and under both pruning strategies.

\begin{table*}[!t]
\centering
\caption{Comparison of fine-tuning methods on sparse LLaMA models at sparsity level $\rho = 0.6$. Results report average and per-task zero-shot accuracy on seven commonsense reasoning tasks. Higher values indicate better performance.}
\label{tab:merge}
\resizebox{\linewidth}{!}{
\begin{tabular}{clccccccccc}
\toprule
\textbf{LLaMA} & \textbf{Method}  & \textbf{WG} & \textbf{RTE} & \textbf{OBQA} & \textbf{HS} & \textbf{BoolQ}  & \textbf{ARC-e} & \textbf{ARC-c} &  \textbf{Avg} \\
\midrule
\multirow{7}{*}{V2-7B} 
& Wanda+LoRA$^*$  & 73.48& 54.51& 29.00 & 51.49& 70.98 & 72.81& 39.07 & 55.91 \\
& Wanda+SPP        & 73.79 & 54.15 & 27.40 & 51.97 & 71.80 & 70.54 & 37.63 & 55.32  \\
& Wanda+SQFT &75.61& 54.87& 29.00& 52.86& 70.46& 72.51& 39.07 & 56.34\\
& Wanda+SEFT        &75.85& 56.32& 29.80& 52.34& 67.65& 72.60& 41.21 & \textbf{56.54}  \\
\cmidrule{2-10}
& SparseGPT+LoRA$^*$  & 73.16& 53.79& 28.60& 51.13& 77.73& 72.34 & 38.48& 56.46 \\
& SparseGPT+SPP        & 72.69& 58.48& 28.80& 52.63& 70.55& 69.91 & 39.16& 56.03  \\
& SparseGPT+SQFT       & 75.37 & 53.43 & 30.00 & 52.38 & 74.95 & 72.89 & 41.38 & 57.20\\
& SparseGPT+SEFT        & 75.53& 55.23& 30.80& 52.99& 77.98& 72.64 & 41.13& \textbf{58.04} \\
\midrule
\multirow{6}{*}{V3-8B} 
& Wanda+LoRA$^*$  & 73.87& 56.68& 27.20& 50.69& 69.91& 71.08 & 39.51& 55.56\\
& Wanda+SPP        & 73.32& 54.51& 27.40& 50.37& 70.46& 70.75 & 37.97& 54.97 \\
& Wanda+SQFT       & 75.84 & 60.26 & 26.40 & 50.85 & 77.61 & 70.24 & 40.01 & 57.32\\
& Wanda+SEFT        & 76.87& 57.76& 28.60& 52.20& 80.83& 71.46 & 41.38& \textbf{58.44} \\
\cmidrule{2-10}
& SparseGPT+LoRA$^*$  & 79.48& 57.04& 29.40& 53.78& 83.30& 74.20 & 43.34& 60.07\\
& SparseGPT+SPP        & 76.87& 57.76& 31.20& 53.45& 80.67& 74.12 & 43.09 & 59.60 \\
& SparseGPT+SQFT  & 79.43& 63.17& 31.40 & 53.83& 82.09 & 73.19& 41.97 & 60.73 \\
& SparseGPT+SEFT        & 78.37& 62.82& 31.20& 53.99& 83.00& 73.40 & 43.69& \textbf{60.92}\\
\midrule
\multirow{6}{*}{V1-13B} 
& Wanda+LoRA$^*$  & 77.11& 59.93& 33.60& 56.37& 82.38& 76.56 & 44.88& 61.54 \\
& Wanda+SPP        & 80.03& 62.09& 32.60& 57.08& 76.73& 75.88 & 44.71& 61.30 \\
& Wanda+SQFT       &79.48 & 58.84 & 32.40 & 56.84 & 82.54 & 76.05 & 46.42 & 61.79\\
& Wanda+SEFT        & 79.32& 59.20& 33.00& 56.58& 83.58& 77.02 & 47.44& \textbf{62.31} \\
\cmidrule{2-10}
& SparseGPT+LoRA$^*$  & 78.37& 59.93& 31.20& 55.23& 79.60& 76.01 & 43.26& 60.51\\
& SparseGPT+SPP        & 80.66& 56.68& 35.00& 57.53& 79.78& 75.76 & 45.22& 61.52  \\
& SparseGPT+SQFT      & 79.63 & 57.28 & 34.20 & 57.50 & 83.45 & 76.01 & 44.79 & 61.84 \\
& SparseGPT+SEFT        & 80.19& 57.40& 33.20& 56.81& 83.52& 77.02 & 45.90& \textbf{62.01} \\

\bottomrule
\end{tabular}
}
\vskip -0.1in
\end{table*}

\subsubsection{Supervised fine-tuning} 
In this section, we evaluate the effectiveness of SEFT on a widely used supervised fine-tuning tasks-Commonsense Reasoning.
Table~\ref{tab:merge} presents the zero-shot accuracy on seven tasks from the commonsense reasoning benchmark. 
To begin, the evaluation is performed on post-training sparse models generated by Wanda and SparseGPT at a sparsity level of 60\%, which provides a balanced trade-off between efficiency and decent performance comparison on this benchmark. The sparse models are then fine-tuned using SEFT and other sparsity-preserving baselines.
We include results for three model sizes: 7B, 8B, and 13B, reporting both per task and average performance.

The results show that SEFT consistently outperforms existing sparsity-preserved fine-tuning methods such as SPP and SQFT across all evaluated models and tasks. SEFT achieves competitive or higher average accuracy in most settings. In particular, SEFT achieves particularly strong results in OBQA and ARC-c over other baselines.
Overall, these findings highlight SEFT's effectiveness in enhancing the overall performance across commonsense reasoning tasks and model sizes.

\subsubsection{Memory and computation efficiency}
\label{main_speedup}

\begin{figure}[!t]
    \centering
    \includegraphics[width=\textwidth]{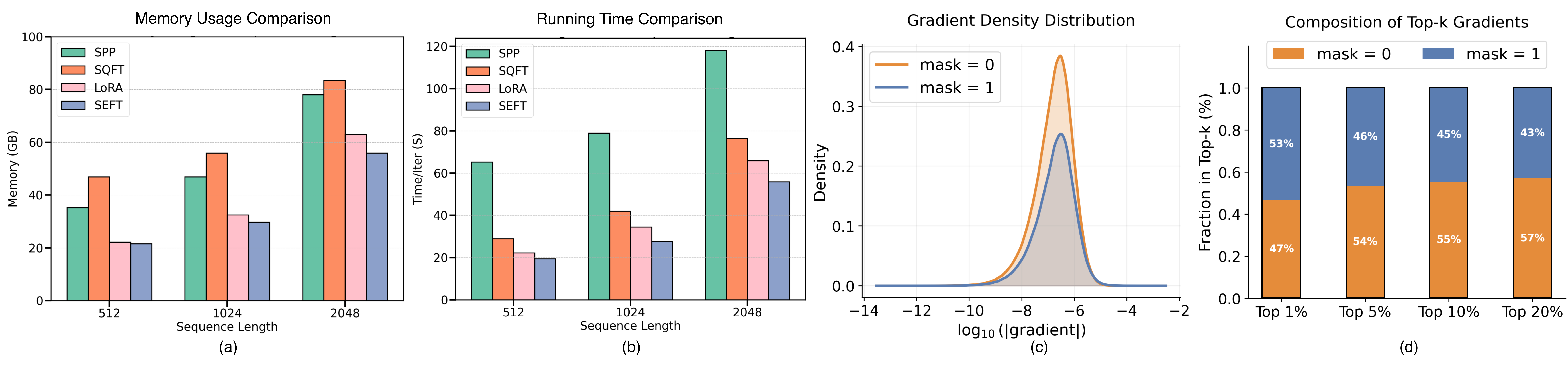}
    \vskip -0.1in
    \caption{(a) GPU memory usage (in GB) and (b) average training time per step (in seconds) for SEFT and baseline methods on LLaMA2-7B using an H100 GPU. All results are reported without activation checkpointing. (c) and (d) are gradient analysis on LLaMA2-7B pruned with Wanda, comparing pruned (\texttt{mask=0}) and retained (\texttt{mask=1}) positions in terms of gradient distribution and top-k composition.}
    \vskip -0.1in
    \label{fig: speed}
\end{figure}

\textbf{Memory Efficiency.} In this section, we evaluate the memory overhead of different sparsity-preserving fine-tuning methods using the LLaMA2-7B model. 
Experiments were conducted with sequence lengths ranging from 512 to 2048 at a rank of 32. As shown in Figure~\ref{fig: speed}~(a), memory usage increases with sequence length for all methods, but SEFT consistently requires less memory than the alternatives, in some cases using about half the memory of SQFT.

Compared with LoRA methods, SEFT exhibits slightly lower memory usage as sequence length increases. As shown in the memory consumption breakdown in Appendix~\ref{sec:mem_usage}, most of these savings come from reduced \emph{activation memory} during fine-tuning. Unlike LoRA-style adapters, which introduce additional activation tensors, SEFT directly updates parameters without auxiliary activations. These results highlight SEFT’s favorable memory-efficiency profile, offering a good balance between resource usage and performance when fine-tuning sparse LLMs.

\textbf{Computation Efficiency.}  We evaluate both training and inference efficiency of SEFT. 
For training, Figure~\ref{fig: speed}~(b) shows that SEFT requires less training time than other sparsity-preserving fine-tuning methods across different sequence lengths. 
For inference, Table~\ref{tab:speed} in Appendix~\ref{sec:speed} reports end-to-end decoding latency of the SEFT fine-tuned sparse LLM using the DeepSparse inference engine~\citep{deepsparse} on an Intel Xeon Platinum 8360Y CPU with 36 cores. The results show that sparse LLMs achieve substantial speedups over their dense counterparts, reaching up to 2.5× improvement at 70\% sparsity. Moreover, the speedup becomes even more pronounced at higher sparsity levels, demonstrating the potential of sparsity-preserving to enable practical deployment benefits.

\subsection{Ablation Study}



To better understand the contribution of each component in SEFT, we conduct two ablation studies: \textit{SEFT w/o OOM}, which disables out-of-mask exploration and restricts sparse updates to the originally retained weights only, and \textit{SEFT w/o DSU}, which removes Delta Support Update and keeps the delta support fixed during fine-tuning.

\begin{wraptable}{r}{0.55\textwidth}
\vskip -0.2in
\begin{minipage}{0.55\textwidth}
\centering
\caption{Performance evaluation of LLaMA1-7B and LLaMA3-8B models on LM-eval under different N:M sparsity patterns. Higher values indicate better performance.}
\label{tab:ablation}
\resizebox{\linewidth}{!}{
\begin{tabular}{l|cc|cc}
\toprule[1pt]
\multirow{2}{*}{\textbf{Method}} & \multicolumn{2}{c|}{\textbf{LLaMA2-7B}} & \multicolumn{2}{c}{\textbf{LLaMA3-8B}} \\ [1pt] 
\cline{2-5} 
\rule{0pt}{2.5ex} &$\rho=0.6$& $\rho=0.7$ & $\rho=0.6$ & $\rho=0.7$ \\
\cmidrule{1-5}
SEFT & \textbf{56.54} & \textbf{48.87} & \textbf{58.44} & \textbf{47.44} \\
\textit{w/o OOM} & 55.98 & 48.36 & 58.13 & 47.02\\
\textit{w/o DSU} & 54.16 & 46.77 & 56.55 & 45.81\\

\bottomrule[1pt]
\end{tabular}
}
\vskip -0.2in
\end{minipage}
\end{wraptable}

Table~\ref{tab:ablation} reports the results on LLaMA2-7B and LLaMA3-8B under 60\% and 70\% sparsity. SEFT consistently achieves the best performance across all settings. Compared with \textit{SEFT w/o DSU}, the performance drop is more pronounced, showing that fixing the delta support leads to clear degradation on both model families and at both sparsity levels. Compared with \textit{SEFT w/o OOM}, the full SEFT also achieves consistently better results, indicating that allowing updates to reach previously pruned weights is beneficial for downstream adaptation. 
This observation is further supported by Figure~\ref{fig: speed} (c) and (d), where pruned weights (\texttt{mask=0}) occupy a large proportion of the top-gradient positions. This indicates that many important task-relevant updates lie outside the retained sparse structure, which directly motivates the design of SEFT to explore and reactivate previously pruned weights.

In Section \ref{sec:adaptation}, we proposed a \textit{Sparsity Topology Adaptation} process to restore the model to the desired sparsity level, as sparse topology evolution in SEFT without a mask constraint may result in a denser model by reactivating previously inactive weights.
We evaluate its effect on LLaMA1-7B, LLaMA2-7B, and LLaMA3-8B pruned to 60\% sparsity with Wanda, by comparing SEFT with and without this component. As shown in Appendix ~\ref{sec:adapt} Table~\ref{tab:maskz_zeros1}, the results indicate that without sparsity topology adaptation, the sparse LLMs become slightly denser after fine-tuning (about 59.6\% sparsity). In contrast, with this component, the models maintain their original sparsity levels while achieving comparable performance.

\subsection{Effect on Other Tasks and Architectures}

To further evaluate the generality of SEFT across tasks and model architectures, we conduct experiments on two challenging benchmarks, MMLU and GSM8K, and extend our study to additional model backbones, including DeepSeek-7B-Chat~\citep{bi2024deepseek} and Mistral-7B-v0.1~\citep{jiang2023mistral7b}. MMLU covers a broad range of academic and professional subjects and evaluates both factual knowledge and reasoning ability, while GSM8K focuses on grade-school math word problems.

\begin{wraptable}{r}{0.55\textwidth}
\vskip -0.1in
\begin{minipage}{0.55\textwidth}
\centering
\caption{Performance evaluation of DeepSeek-7B-chat and Mistral-7B-v0.1 models on MMLU and GSM8K benchmark at a sparsity level of 60\% by Wanda pruning method. Higher values indicate better performance.}
\label{tab:others}
\resizebox{\linewidth}{!}{
\begin{tabular}{ll|c|c}
\toprule[1pt]
{\textbf{Method}} & & {\textbf{MMLU}} & {\textbf{GSM8K}} \\ 
\cmidrule{1-4}
\multirow{4}{*}{\textbf{DeepSeek-7B-chat}} 
& LoRA* & 45.57 &31.46 \\
& SPP & 45.74 & 31.45\\
& SQFT & 44.23 & 31.97\\ 
& SEFT & \textbf{46.14} & \textbf{32.52}\\ 
\cmidrule{1-4}
\multirow{4}{*}{\textbf{Mistral-7B-v0.1}} 
& LoRA* &53.52 & 37.83\\
& SPP & 53.78  & 39.73\\
& SQFT & 53.57 & 40.86\\ 
& SEFT & \textbf{53.97} & \textbf{42.91}\\
\bottomrule[1pt]
\end{tabular}
}
\end{minipage}
\vskip -0.1in
\end{wraptable}

Specifically, we evaluate fine-tuning on sparse models, using DeepSeek-7B-Chat and Mistral-7B-v0.1 pruned by Wanda to 60\% sparsity. We compare SEFT with LoRA*, SPP, and SQFT, and evaluate all methods on both MMLU and GSM8K. The results show that SEFT consistently outperforms the competing methods across both tasks and architectures. In particular, SEFT achieves strong performance on MMLU and up to around 2\% improvement on GSM8K on Mistral-7B-v0.1. These findings demonstrate the robustness of SEFT on diverse tasks and its ability to generalize across different model backbones, highlighting its potential as both task- and model-agnostic fine-tuning method.

\subsection{Impact of Sparsity Level and Pattern}

\begin{wraptable}{r}{0.5\textwidth}
\vskip -0.2in
\begin{minipage}{0.5\textwidth}
\centering
\centering
\caption{Performance comparison of LLaMA2-7B on LM-eval under different sparsity levels ($\rho$).}
\label{tab:sparsity}
\resizebox{\linewidth}{!}{
\begin{tabular}{l|c|c|c|c}
\toprule[1pt]
\textbf{Method} & $\rho$=0.5 & $\rho$=0.6 & $\rho$=0.7 & $\rho$=0.8 \\ 
\cmidrule{1-5}
LoRA* & 60.19& 56.98& 49.07& 39.02 \\
SPP & 59.92& 56.56& 49.63& 40.40 \\
SQFT & 60.37& 57.84& 51.58& 40.70 \\
SEFT & \textbf{60.94}& \textbf{58.50}& \textbf{52.74}& \textbf{42.06} \\ 
\bottomrule[1pt]
\end{tabular}
}
\vskip -0.2in
\end{minipage}
\end{wraptable}

We further evaluate the impact of sparsity levels for SEFT. Table~\ref{tab:sparsity} reports LM-eval performance of LLaMA2-7B with sparsity levels $\rho \in \{0.5, 0.6, 0.7, 0.8\}$. SEFT achieves better results across sparsity levels, and its advantage grows as $\rho$ increases. At $\rho=0.8$, for instance, SEFT outperforms the strongest baseline by 1.36 points. The same trend is observed on additional tasks and model scales (Appendix~\ref{sec:sparsity}, Table~\ref{tab:sparsity_level}), suggesting that allowing sparse updateS to evolve during fine-tuning is particularly beneficial in high-sparsity regimes.

We also evaluate SEFT under N:M sparsity pattern, which is increasingly supported by modern hardware \citep{nvidia2021ampere}. To satisfy the N:M constraint, Delta Support Update is restricted to the currently active weights (\texttt{mask=1}).
As shown in Appendix~\ref{sec:nm} Table~\ref{tab:nm}, SEFT remains effective under 2:4 and 4:8 sparsity on LLaMA2-7B and LLaMA3-8B, consistently outperforming other sparsity-preserving baselines. These results show that SEFT extends naturally beyond unstructured pruning to hardware-friendly semi-structured sparsity settings.

Finally, we study the impact of several SEFT design choices, including (i) the support refresh ratio controlled by $\tau(t)$ (Appendix~\ref{sec:drop}), (ii) the update frequency of Delta Support Update (Appendix~\ref{sec:freq}), and (iii) the number of fine-tuning parameters (Appendix~\ref{sec:ranks}).

\section{Conclusion}

In this work, we propose Sparsity Evolution Fine-Tuning (SEFT), a fine-tuning framework for adapting sparse large language models by allowing sparse structure to evolve with downstream tasks. SEFT combines Delta Support Update, which refreshes where sparse updates are allocated, with Sparse Topology Adaptation, which preserves the target sparsity level of the model. In this way, SEFT enables more flexible adaptation of sparse LLMs while maintaining the efficiency advantages of sparsity.

Extensive experiments across multiple model families, pruning settings, and downstream tasks show that SEFT improves adaptation quality over existing baselines. We further show that these gains come with favorable memory usage and training efficiency, and that SEFT remains effective across different sparsity levels and structured sparsity patterns. Overall, our results highlight SEFT as a practical and effective framework for fine-tuning sparse LLMs.

We include related work in Appendix~\ref{sec:related_work}, and discuss limitations and future directions in Appendix~\ref{app:limits}.

\section*{Acknowledgements}
This work is part of the research program ‘MegaMind - Measuring, Gathering, Mining and Integrating Data for Self-management in the Edge of the Electricity System’, (partly) financed by the Dutch Research Council (NWO) through the Perspectief program under number P19-25. This work is partly supported by the SmartCHANGE project, funded within EU’s Horizon Europe research program (GA No. 101080965).
This work used the Dutch national e-infrastructure with the support of the SURF Cooperative, using grant No. EINF-12291 and NWO-2023.027.

\bibliographystyle{plain}  
\bibliography{example_paper}

@article{ansell2024scaling,
  title={Scaling sparse fine-tuning to large language models},
  author={Ansell, Alan and Vuli{\'c}, Ivan and Sterz, Hannah and Korhonen, Anna and Ponti, Edoardo M},
  journal={arXiv preprint arXiv:2401.16405},
  year={2024}
}

@inproceedings{frantar2023sparsegpt,
  title={Sparsegpt: Massive language models can be accurately pruned in one-shot},
  author={Frantar, Elias and Alistarh, Dan},
  booktitle={International Conference on Machine Learning},
  pages={10323--10337},
  year={2023},
  organization={PMLR}
}

@article{sun2023simple,
  title={A simple and effective pruning approach for large language models},
  author={Sun, Mingjie and Liu, Zhuang and Bair, Anna and Kolter, J Zico},
  journal={arXiv preprint arXiv:2306.11695},
  year={2023}
}

@inproceedings{yinoutlier,
  title={Outlier Weighed Layerwise Sparsity (OWL): A Missing Secret Sauce for Pruning LLMs to High Sparsity},
  author={Yin, Lu and Wu, You and Zhang, Zhenyu and Hsieh, Cheng-Yu and Wang, Yaqing and Jia, Yiling and Li, Gen and JAISWAL, AJAY KUMAR and Pechenizkiy, Mykola and Liang, Yi and others},
  booktitle={International Conference on Machine Learning},
  year={2023},
  organization={PMLR}
}

@article{ma2023llm,
  title={Llm-pruner: On the structural pruning of large language models},
  author={Ma, Xinyin and Fang, Gongfan and Wang, Xinchao},
  journal={Advances in neural information processing systems},
  volume={36},
  pages={21702--21720},
  year={2023}
}

@article{naveed2023comprehensive,
  title={A comprehensive overview of large language models},
  author={Naveed, Humza and Khan, Asad Ullah and Qiu, Shi and Saqib, Muhammad and Anwar, Saeed and Usman, Muhammad and Akhtar, Naveed and Barnes, Nick and Mian, Ajmal},
  journal={arXiv preprint arXiv:2307.06435},
  year={2023}
}

@article{zhu2024survey,
  title={A survey on model compression for large language models},
  author={Zhu, Xunyu and Li, Jian and Liu, Yong and Ma, Can and Wang, Weiping},
  journal={Transactions of the Association for Computational Linguistics},
  volume={12},
  pages={1556--1577},
  year={2024},
  publisher={MIT Press 255 Main Street, 9th Floor, Cambridge, Massachusetts 02142, USA~…}
}

@article{hu2021lora,
  title={Lora: Low-rank adaptation of large language models},
  author={Hu, Edward J and Shen, Yelong and Wallis, Phillip and Allen-Zhu, Zeyuan and Li, Yuanzhi and Wang, Shean and Wang, Lu and Chen, Weizhu},
  journal={arXiv preprint arXiv:2106.09685},
  year={2021}
}

@article{mangrulkar2022peft,
  title={Peft: State-of-the-art parameter-efficient fine-tuning methods},
  author={Mangrulkar, Sourab and Gugger, Sylvain and Debut, Lysandre and Belkada, Younes and Paul, Sayak and Bossan, B},
  journal={URL: https://github. com/huggingface/peft},
  year={2022}
}

@article{xu2023qa,
  title={Qa-lora: Quantization-aware low-rank adaptation of large language models},
  author={Xu, Yuhui and Xie, Lingxi and Gu, Xiaotao and Chen, Xin and Chang, Heng and Zhang, Hengheng and Chen, Zhengsu and Zhang, Xiaopeng and Tian, Qi},
  journal={arXiv preprint arXiv:2309.14717},
  year={2023}
}

@article{dettmers2024qlora,
  title={Qlora: Efficient finetuning of quantized llms},
  author={Dettmers, Tim and Pagnoni, Artidoro and Holtzman, Ari and Zettlemoyer, Luke},
  journal={Advances in Neural Information Processing Systems},
  volume={36},
  year={2024}
}

@article{zhao2024galore,
  title={Galore: Memory-efficient llm training by gradient low-rank projection},
  author={Zhao, Jiawei and Zhang, Zhenyu and Chen, Beidi and Wang, Zhangyang and Anandkumar, Anima and Tian, Yuandong},
  journal={arXiv preprint arXiv:2403.03507},
  year={2024}
}

@article{ding2023parameter,
  title={Parameter-efficient fine-tuning of large-scale pre-trained language models},
  author={Ding, Ning and Qin, Yujia and Yang, Guang and Wei, Fuchao and Yang, Zonghan and Su, Yusheng and Hu, Shengding and Chen, Yulin and Chan, Chi-Min and Chen, Weize and others},
  journal={Nature Machine Intelligence},
  volume={5},
  number={3},
  pages={220--235},
  year={2023},
  publisher={Nature Publishing Group UK London}
}

@article{han2024parameter,
  title={Parameter-efficient fine-tuning for large models: A comprehensive survey},
  author={Han, Zeyu and Gao, Chao and Liu, Jinyang and Zhang, Jeff and Zhang, Sai Qian},
  journal={arXiv preprint arXiv:2403.14608},
  year={2024}
}

@article{lu2024alphapruning,
  title={Alphapruning: Using heavy-tailed self regularization theory for improved layer-wise pruning of large language models},
  author={Lu, Haiquan and Zhou, Yefan and Liu, Shiwei and Wang, Zhangyang and Mahoney, Michael W and Yang, Yaoqing},
  journal={Advances in Neural Information Processing Systems},
  volume={37},
  pages={9117--9152},
  year={2024}
}

@inproceedings{luspp,
  title={SPP: Sparsity-Preserved Parameter-Efficient Fine-Tuning for Large Language Models},
  author={Lu, Xudong and Zhou, Aojun and Xu, Yuhui and Zhang, Renrui and Gao, Peng and Li, Hongsheng},
  booktitle={International Conference on Machine Learning},
  year={2024},
  organization={PMLR}
}

@inproceedings{munoz2024sqft,
  title={SQFT: Low-cost Model Adaptation in Low-precision Sparse Foundation Models},
  author={Munoz, Juan and Yuan, Jinjie and Jain, Nilesh},
  booktitle={Findings of the Association for Computational Linguistics: EMNLP 2024},
  pages={12817--12832},
  year={2024}
}

@article{mocanu2018scalable,
  title={Scalable training of artificial neural networks with adaptive sparse connectivity inspired by network science},
  author={Mocanu, Decebal Constantin and Mocanu, Elena and Stone, Peter and Nguyen, Phuong H and Gibescu, Madeleine and Liotta, Antonio},
  journal={Nature communications},
  volume={9},
  number={1},
  pages={2383},
  year={2018},
  publisher={Nature Publishing Group UK London}
}

@inproceedings{graesser2022state,
  title={The State of Sparse Training in Deep Reinforcement Learning},
  author={Graesser, Laura and Evci, Utku and Elsen, Erich and Castro, Pablo Samuel},
  booktitle={International Conference on Machine Learning},
  pages={7766--7792},
  year={2022},
  organization={PMLR}
}

@inproceedings{xiaodynamic,
  title={Dynamic Sparse Network for Time Series Classification: Learning What to “See”},
  author={Xiao, Qiao and Wu, Boqian and Zhang, Yu and Liu, Shiwei and Pechenizkiy, Mykola and Mocanu, Elena and Mocanu, Decebal Constantin},
  booktitle={Advances in Neural Information Processing Systems},
    year={2022}
}

@inproceedings{evci2020rigging,
  title={Rigging the lottery: Making all tickets winners},
  author={Evci, Utku and Gale, Trevor and Menick, Jacob and Castro, Pablo Samuel and Elsen, Erich},
  booktitle={International Conference on Machine Learning},
  pages={2943--2952},
  year={2020},
  organization={PMLR}
}

@article{yuan2021mest,
  title={Mest: Accurate and fast memory-economic sparse training framework on the edge},
  author={Yuan, Geng and Ma, Xiaolong and Niu, Wei and Li, Zhengang and Kong, Zhenglun and Liu, Ning and Gong, Yifan and Zhan, Zheng and He, Chaoyang and Jin, Qing and others},
  journal={Advances in Neural Information Processing Systems},
  volume={34},
  pages={20838--20850},
  year={2021}
}

@inproceedings{mostafa2019parameter,
  title={Parameter efficient training of deep convolutional neural networks by dynamic sparse reparameterization},
  author={Mostafa, Hesham and Wang, Xin},
  booktitle={International Conference on Machine Learning},
  pages={4646--4655},
  year={2019},
  organization={PMLR}
}

@misc{
dettmers2020sparse,
title={Sparse Networks from Scratch: Faster Training without Losing Performance},
author={Tim Dettmers and Luke Zettlemoyer},
year={2020}}

@article{atashgahi2022quick,
  title={Quick and robust feature selection: the strength of energy-efficient sparse training for autoencoders},
  author={Atashgahi, Zahra and Sokar, Ghada and van der Lee, Tim and Mocanu, Elena and Mocanu, Decebal Constantin and Veldhuis, Raymond and Pechenizkiy, Mykola},
  journal={Machine Learning},
  pages={1--38},
  year={2022},
  publisher={Springer}
}

@article{sokar2022pay,
  title={Where to pay attention in sparse training for feature selection?},
  author={Sokar, Ghada and Atashgahi, Zahra and Pechenizkiy, Mykola and Mocanu, Decebal Constantin},
  journal={Advances in Neural Information Processing Systems},
  volume={35},
  pages={1627--1642},
  year={2022}
}

@article{zhu2023minigpt,
  title={Minigpt-4: Enhancing vision-language understanding with advanced large language models},
  author={Zhu, Deyao and Chen, Jun and Shen, Xiaoqian and Li, Xiang and Elhoseiny, Mohamed},
  journal={arXiv preprint arXiv:2304.10592},
  year={2023}
}

@article{wei2022chain,
  title={Chain-of-thought prompting elicits reasoning in large language models},
  author={Wei, Jason and Wang, Xuezhi and Schuurmans, Dale and Bosma, Maarten and Xia, Fei and Chi, Ed and Le, Quoc V and Zhou, Denny and others},
  journal={Advances in neural information processing systems},
  volume={35},
  pages={24824--24837},
  year={2022}
}

@inproceedings{vaithilingam2022expectation,
  title={Expectation vs. experience: Evaluating the usability of code generation tools powered by large language models},
  author={Vaithilingam, Priyan and Zhang, Tianyi and Glassman, Elena L},
  booktitle={Chi conference on human factors in computing systems extended abstracts},
  pages={1--7},
  year={2022}
}

@article{hu2025lors,
  title={LoRS: Efficient Low-Rank Adaptation for Sparse Large Language Model},
  author={Hu, Yuxuan and Zhang, Jing and Chen, Xiaodong and Zhao, Zhe and Li, Cuiping and Chen, Hong},
  journal={arXiv preprint arXiv:2501.08582},
  year={2025}
}

@article{touvron2023llama2,
  title={Llama 2: Open foundation and fine-tuned chat models},
  author={Touvron, Hugo and Martin, Louis and Stone, Kevin and Albert, Peter and Almahairi, Amjad and Babaei, Yasmine and Bashlykov, Nikolay and Batra, Soumya and Bhargava, Prajjwal and Bhosale, Shruti and others},
  journal={arXiv preprint arXiv:2307.09288},
  year={2023}
}

@article{dubey2024llama3,
  title={The llama 3 herd of models},
  author={Dubey, Abhimanyu and Jauhri, Abhinav and Pandey, Abhinav and Kadian, Abhishek and Al-Dahle, Ahmad and Letman, Aiesha and Mathur, Akhil and Schelten, Alan and Yang, Amy and Fan, Angela and others},
  journal={arXiv preprint arXiv:2407.21783},
  year={2024}
}

@article{touvron2023llama1,
  title={Llama: Open and efficient foundation language models},
  author={Touvron, Hugo and Lavril, Thibaut and Izacard, Gautier and Martinet, Xavier and Lachaux, Marie-Anne and Lacroix, Timoth{\'e}e and Rozi{\`e}re, Baptiste and Goyal, Naman and Hambro, Eric and Azhar, Faisal and others},
  journal={arXiv preprint arXiv:2302.13971},
  year={2023}
}

@article{merity2016pointer,
  title={Pointer sentinel mixture models},
  author={Merity, Stephen and Xiong, Caiming and Bradbury, James and Socher, Richard},
  journal={arXiv preprint arXiv:1609.07843},
  year={2016}
}

@article{raffel2020exploring,
  title={Exploring the limits of transfer learning with a unified text-to-text transformer},
  author={Raffel, Colin and Shazeer, Noam and Roberts, Adam and Lee, Katherine and Narang, Sharan and Matena, Michael and Zhou, Yanqi and Li, Wei and Liu, Peter J},
  journal={Journal of machine learning research},
  volume={21},
  number={140},
  pages={1--67},
  year={2020}
}

@inproceedings{gale2020sparse,
  title={Sparse gpu kernels for deep learning},
  author={Gale, Trevor and Zaharia, Matei and Young, Cliff and Elsen, Erich},
  booktitle={SC20: International Conference for High Performance Computing, Networking, Storage and Analysis},
  pages={1--14},
  year={2020},
  organization={IEEE}
}

@article{gao2021framework,
  title={A framework for few-shot language model evaluation},
  author={Gao, Leo and Tow, Jonathan and Biderman, Stella and Black, Sid and DiPofi, Anthony and Foster, Charles and Golding, Laurence and Hsu, Jeffrey and McDonell, Kyle and Muennighoff, Niklas and others},
  journal={Version v0. 0.1. Sept},
  volume={10},
  pages={8--9},
  year={2021}
}

@article{hu2023llm,
  title={Llm-adapters: An adapter family for parameter-efficient fine-tuning of large language models},
  author={Hu, Zhiqiang and Wang, Lei and Lan, Yihuai and Xu, Wanyu and Lim, Ee-Peng and Bing, Lidong and Xu, Xing and Poria, Soujanya and Lee, Roy Ka-Wei},
  journal={arXiv preprint arXiv:2304.01933},
  year={2023}
}

@article{hendrycks2020measuring,
  title={Measuring massive multitask language understanding},
  author={Hendrycks, Dan and Burns, Collin and Basart, Steven and Zou, Andy and Mazeika, Mantas and Song, Dawn and Steinhardt, Jacob},
  journal={arXiv preprint arXiv:2009.03300},
  year={2020}
}

@inproceedings{lee2019snip,
  title={SNIP: single-shot network pruning based on connection sensitivity},
  author={Lee, N and Ajanthan, T and Torr, P},
  booktitle={International Conference on Learning Representations},
  year={2019},
  organization={Open Review}
}

@article{nowak2024fantastic,
  title={Fantastic weights and how to find them: Where to prune in dynamic sparse training},
  author={Nowak, Aleksandra and Grooten, Bram and Mocanu, Decebal Constantin and Tabor, Jacek},
  journal={Advances in Neural Information Processing Systems},
  volume={36},
  year={2024}
}

@inproceedings{wu2023bold,
  title={Bold but cautious: Unlocking the potential of personalized federated learning through cautiously aggressive collaboration},
  author={Wu, Xinghao and Liu, Xuefeng and Niu, Jianwei and Zhu, Guogang and Tang, Shaojie},
  booktitle={Proceedings of the IEEE/CVF International Conference on Computer Vision},
  pages={19375--19384},
  year={2023}
}

@inproceedings{guo-etal-2021-parameter,
    title = "Parameter-Efficient Transfer Learning with Diff Pruning",
    author = "Guo, Demi  and
      Rush, Alexander  and
      Kim, Yoon",
    booktitle = "Proceedings of the 59th Annual Meeting of the Association for Computational Linguistics and the 11th International Joint Conference on Natural Language Processing (Volume 1: Long Papers)",
    month = aug,
    year = "2021",
    pages = "4884--4896",
}

@inproceedings{xu-etal-2021-raise,
    title = "Raise a Child in Large Language Model: Towards Effective and Generalizable Fine-tuning",
    author = "Xu, Runxin  and
      Luo, Fuli  and
      Zhang, Zhiyuan  and
      Tan, Chuanqi  and
      Chang, Baobao  and
      Huang, Songfang  and
      Huang, Fei",
    booktitle = "Proceedings of the 2021 Conference on Empirical Methods in Natural Language Processing",
    month = nov,
    year = "2021",
    pages = "9514--9528",
}

@inproceedings{ansell-etal-2022-composable,
    title = "Composable Sparse Fine-Tuning for Cross-Lingual Transfer",
    author = "Ansell, Alan  and
      Ponti, Edoardo  and
      Korhonen, Anna  and
      Vuli{\'c}, Ivan",
    booktitle = "Proceedings of the 60th Annual Meeting of the Association for Computational Linguistics (Volume 1: Long Papers)",
    month = may,
    year = "2022",
    pages = "1778--1796",
}

@article{
pfeiffer2023modular,
title={Modular Deep Learning},
author={Jonas Pfeiffer and Sebastian Ruder and Ivan Vuli{\'c} and Edoardo Ponti},
journal={Transactions on Machine Learning Research},
issn={2835-8856},
year={2023},
url={https://openreview.net/forum?id=z9EkXfvxta},
note={Survey Certification}
}

@article{
deepsparse,
title={NeuralMagic DeepSparse Inference Engine},
author={DeepSparse},
year={2021},
url={https://openreview.net/forum?id=z9EkXfvxta},
}

@inproceedings{LiuCCCXWKPMW23,
  author={Shiwei Liu and Tianlong Chen and Xiaohan Chen and Xuxi Chen and Qiao Xiao and Boqian Wu and Tommi Kärkkäinen and Mykola Pechenizkiy and Decebal Constantin Mocanu and Zhangyang Wang},
  title={More ConvNets in the 2020s: Scaling up Kernels Beyond 51x51 using Sparsity},
  year={2023},
  url={https://openreview.net/pdf?id=bXNl-myZkJl},
  booktitle={ICLR},
}

@inproceedings{
wu2025dynamic,
title={Dynamic Sparse Training versus Dense Training: The Unexpected Winner in Image Corruption Robustness},
author={Boqian Wu and Qiao Xiao and Shunxin Wang and Nicola Strisciuglio and Mykola Pechenizkiy and Maurice van Keulen and Decebal Constantin Mocanu and Elena Mocanu},
booktitle={The Thirteenth International Conference on Learning Representations},
year={2025},
url={https://openreview.net/forum?id=daUQ7vmGap}
}

@inproceedings{
wu2024eenet,
title={E2{EN}et: Dynamic Sparse Feature Fusion for Accurate and Efficient 3D Medical Image Segmentation},
author={Boqian Wu and Qiao Xiao and Shiwei Liu and Lu Yin and Mykola Pechenizkiy and Decebal Constantin Mocanu and Maurice van Keulen and Elena Mocanu},
booktitle={The Thirty-eighth Annual Conference on Neural Information Processing Systems},
year={2024},
url={https://openreview.net/forum?id=Xp8qhdmeb4}
}

@article{cobbe2021training,
  title={Training verifiers to solve math word problems},
  author={Cobbe, Karl and Kosaraju, Vineet and Bavarian, Mohammad and Chen, Mark and Jun, Heewoo and Kaiser, Lukasz and Plappert, Matthias and Tworek, Jerry and Hilton, Jacob and Nakano, Reiichiro and others},
  journal={arXiv preprint arXiv:2110.14168},
  year={2021}
}

@misc{jiang2023mistral7b,
      title={Mistral 7B}, 
      author={Albert Q. Jiang and Alexandre Sablayrolles and Arthur Mensch and Chris Bamford and Devendra Singh Chaplot and Diego de las Casas and Florian Bressand and Gianna Lengyel and Guillaume Lample and Lucile Saulnier and Lélio Renard Lavaud and Marie-Anne Lachaux and Pierre Stock and Teven Le Scao and Thibaut Lavril and Thomas Wang and Timothée Lacroix and William El Sayed},
      year={2023},
      eprint={2310.06825},
      archivePrefix={arXiv},
      primaryClass={cs.CL},
      url={https://arxiv.org/abs/2310.06825}, 
}

@article{bi2024deepseek,
  title={Deepseek llm: Scaling open-source language models with longtermism},
  author={Bi, Xiao and Chen, Deli and Chen, Guanting and Chen, Shanhuang and Dai, Damai and Deng, Chengqi and Ding, Honghui and Dong, Kai and Du, Qiushi and Fu, Zhe and others},
  journal={arXiv preprint arXiv:2401.02954},
  year={2024}
}

@misc{nvidia2021ampere,
  title        = {NVIDIA Ampere Architecture In-Depth},
  author       = {NVIDIA},
  year         = {2021},
  note         = {\url{https://developer.nvidia.com/blog/nvidia-ampere-architecture-in-depth/}}
}

@article{lie2023cerebras,
  title={Cerebras architecture deep dive: First look inside the hardware/software co-design for deep learning},
  author={Lie, Sean},
  journal={IEEE Micro},
  volume={43},
  number={3},
  pages={18--30},
  year={2023},
  publisher={IEEE}
}

@inproceedings{thangarasa2024sparse,
  title={Sparse-IFT: sparse Iso-FLOP transformations for maximizing training efficiency},
  author={Thangarasa, Vithursan and Saxena, Shreyas and Gupta, Abhay and Lie, Sean},
  booktitle={Proceedings of the 41st International Conference on Machine Learning},
  pages={47997--48018},
  year={2024}
}

@inproceedings{li2024owlore,
  title={OwLore: Outlier-weighed Layerwise Sampled Low-Rank Projection for Memory-Efficient LLM Fine-tuning},
  author={Pengxiang Li and Lu Yin and Xiaowei Gao and Shiwei Liu},
  booktitle={Findings of the Association for Computational Linguistics: ACL 2025}, year={2025}
}

@inproceedings{bandari2024c4,
  title={Is c4 dataset optimal for pruning? an investigation of calibration data for llm pruning},
  author={Bandari, Abhinav and Yin, Lu and Hsieh, Cheng-Yu and Jaiswal, Ajay Kumar and Chen, Tianlong and Shen, Li and Krishna, Ranjay and Liu, Shiwei},
  booktitle={Proceedings of the 2024 Conference on Empirical Methods in Natural Language Processing},
  pages={18089--18099},
  year={2024}
}

@inproceedings{williams2026compressing,
  title={Compressing language models for specialized domains},
  author={Williams, Miles and Chrysostomou, George and Jeronymo, Vitor Amancio and Aletras, Nikolaos},
  booktitle={Proceedings of the 19th Conference of the European Chapter of the Association for Computational Linguistics (Volume 1: Long Papers)},
  pages={7393--7415},
  year={2026}
}

@inproceedings{songsparse,
  title={Sparse is Enough in Fine-tuning Pre-trained Large Language Models},
  author={Song, Weixi and Li, Zuchao and Zhang, Lefei and Du, Bo and others},
  booktitle={Forty-first International Conference on Machine Learning},
  year={2024}
}

@article{bhardwaj2024sparse,
  title={Sparse High Rank Adapters},
  author={Bhardwaj, Kartikeya and Pandey, Nilesh and Priyadarshi, Sweta and Ganapathy, Viswanath and Kadambi, Shreya and Esteves, Rafael and Borse, Shubhankar and Whatmough, Paul and Garrepalli, Risheek and van Baalen, Mart and others},
  journal={Advances in Neural Information Processing Systems},
  volume={37},
  pages={13685--13715},
  year={2024}
}

@inproceedings{LTH,
  title={The lottery ticket hypothesis: Finding sparse, trainable neural networks. },
  author={Jonathan Frankle, Michael Carbin},
  booktitle={ICLR},
  year={2019}
}

@misc{FixedSparseMasks,
      title={Training Neural Networks with Fixed Sparse Masks}, 
      author={Yi-Lin Sung, Varun Nair, Colin Raffel},
      year={2021},
      eprint={2111.09839},
      archivePrefix={arXiv},
      primaryClass={cs.CL},
      url={https://arxiv.org/abs/2111.09839}, 
}

@inproceedings{
he2025smt,
title={{SMT}: Fine-Tuning Large Language Models with Sparse Matrices},
author={Haoze He and Juncheng B Li and Xuan Jiang and Heather Miller},
booktitle={The Thirteenth International Conference on Learning Representations},
year={2025},
url={https://openreview.net/forum?id=GbgCRJedQ7}
}

@inproceedings{zhang2024gradient,
  title={Gradient-based parameter selection for efficient fine-tuning},
  author={Zhang, Zhi and Zhang, Qizhe and Gao, Zijun and Zhang, Renrui and Shutova, Ekaterina and Zhou, Shiji and Zhang, Shanghang},
  booktitle={Proceedings of the IEEE/CVF Conference on Computer Vision and Pattern Recognition},
  pages={28566--28577},
  year={2024}
}

\clearpage

\appendix

\section*{\centering
Appendix}

\subsection*{Table of Contents}
\vspace{0.1cm}
\hrule
\vspace{-0.7cm}

\listofappendix

\vspace{0.2cm}
\hrule
\vspace{0.3cm}

\section{Related Work}
\label{sec:related_work}
\addcontentsline{apc}{section}{Related Work}

\subsection{LLM Pruning}
\addcontentsline{apc}{subsection}{LLM Pruning}

In the era of large language models (LLMs), their enormous sizes pose significant challenges for real-world deployment, including increased computational and memory demands. To address these challenges recent research has increasingly focused on post-training pruning methods, which start from pre-trained networks and remove redundant parameters to reduce model size and complexity \citep{ma2023llm, naveed2023comprehensive, zhu2024survey}. 
For instance, SparseGPT utilizes second-order information to solve a layer-wise reconstruction problem, enabling the effective pruning of large models \citep{frantar2023sparsegpt}. Similarly, Wanda introduces a pruning metric that considers both weight magnitude and corresponding activations~\citep{sun2023simple}. 
More recently, layer-wise prunability has been introduced as a technique to enhance traditional pruning methods by adaptively allocating sparsity across layers based on their individual importance \citep{yinoutlier, lu2024alphapruning}.
{However, existing methods often struggle to maintain satisfactory performance, especially under high sparsity levels. Moreover, state-of-the-art LLM pruning techniques heavily rely on calibration data, which may inadvertently remove connections critical for specific downstream tasks, leading to suboptimal performance. To preserve the efficiency benefits of pruning while maintaining accuracy, we propose SEFT, a fine-tuning method that maintains sparsity throughout training and enables the dynamic evolution and recovery of sparse connectivity. Unlike the method \citep{FixedSparseMasks, LTH} that rely on a dense model and update only a fixed subset of parameters during training, SEFT continually repairs task-relevant connections.}


\subsection{Parameter-efficient Fine-tuning}
\addcontentsline{apc}{subsection}{Parameter-efficient Fine-tuning}

Parameter-Efficient Fine-Tuning (PEFT) has gained significant attention for its ability to adapt large language models (LLMs) to downstream tasks while significantly reducing computational and memory costs \citep{mangrulkar2022peft, ding2023parameter, han2024parameter}. A notable PEFT approach, Low-Rank Adaptation (LoRA) \citep{hu2021lora}, and its variants introduce trainable layer-wise low-rank decomposition matrices into network, allowing efficient parameter updates with minimal overhead \citep{xu2023qa, zhao2024galore, dettmers2024qlora}. {More recent methods attempt to further improve efficiency by selecting a subset of base-model weights for fine-tuning \cite{songsparse}; however, these approaches still require maintaining a dense base model during training.}
However, when fine-tuning sparse LLMs, a challenge that arises is that merging dense adapters with sparse weights leads to the overall loss of sparsity, which negates the efficiency benefits of sparse models. Recent studies such as SPP \citep{luspp} and SQFT \citep{munoz2024sqft} have extended LoRA to support sparse LLMs by incorporating masking mechanisms, thereby preserving sparsity during fine-tuning.


{In contrast, sparse fine-tuning, which updates only a subset of the model’s original parameters with minimal architectural changes, has emerged as a practical strategy for PEFT. However, its practical memory and computational benefits often diminish at large scale. Methods such as SIFT~\citep{songsparse}, SHiRA \citep{bhardwaj2024sparse} and SpIEL \citep{ansell2024scaling} impose sparse updates, yet still rely on dense pretrained models and result in dense models after fine-tuning.
Other approaches, including SMT \citep{he2025smt} and GPS \citep{zhang2024gradient}, improve efficiency by selecting gradients per neuron. Nevertheless, they depend on gradient-based warm-up phases and additional binary masking.
These mechanisms add overhead from mask storage, dense model dependency, or fixing weight selection in advance, making them less suitable for sparse language model fine-tuning across diverse downstream tasks. In contrast, SEFT inherits the advantages of sparsity while dynamically recovering task-relevant connectivity and maintaining an efficient sparse topology throughout training.}


\subsection{Dynamic Sparse Training}
\addcontentsline{apc}{subsection}{Dynamic Sparse Training}

Dynamic Sparse Training (DST) is a paradigm that maintains a small fraction of active parameters throughout training by starting with a sparse neural network and dynamically evolving its sparse connectivity using a prune-and-regrow strategy \citep{mocanu2018scalable, evci2020rigging, LiuCCCXWKPMW23, wu2025dynamic}. It was first well established by \citep{mocanu2018scalable} through the Sparse Evolutionary Training (SET) algorithm, which demonstrated superior performance compared to static sparse neural networks by dynamically evolving sparse connectivity during training. 
Recent advancements have explored diverse pruning criteria, such as magnitude-based \citep{evci2020rigging}, weight-balanced \citep{mocanu2018scalable}, and gradient-based pruning \citep{yuan2021mest}, alongside regrowth strategies guided by randomness \citep{mostafa2019parameter, xiaodynamic}, momentum \citep{dettmers2020sparse}, and gradient information \citep{evci2020rigging}. 
DST has been effectively applied and widely adopted in various domains, including reinforcement learning \citep{graesser2022state}, features selection \citep{sokar2022pay, atashgahi2022quick}, and image segmentation \citep{wu2024eenet}.
Building on this foundation, recent work \citep{ansell2024scaling} was the first to scale DST to fine-tuning LLMs. Our work extends this further to sparse LLMs by introducing dynamic topology evolution and adaptation, enabling efficient and task-specific fine-tuning for downstream applications.  

\section{Inference Speedup on CPU}
\label{sec:speed}
\addcontentsline{apc}{section}{Inference Speedup on CPU}

While support for unstructured sparsity on modern GPUs remains relatively limited, there is growing attention toward enabling such capabilities. For example, Cerebras has developed specialized hardware designed to support unstructured sparsity at scale \citep{lie2023cerebras, thangarasa2024sparse}. Additionally, NVIDIA's Ampere and Hopper architectures have introduced hardware-level support for structured sparsity (e.g., 2:4 patterns), enabling modest acceleration through sparse tensor cores \citep{nvidia2021ampere}. These efforts reflect a broader trend toward making sparsity-aware training and inference feasible on GPU hardware.

Despite these advancements, unstructured sparsity has shown immediate and practical benefits on non-GPU platforms such as CPUs and custom accelerators. For instance, FPGA-based accelerators for sparse RNNs have achieved notable gains in speed and energy efficiency by fully utilizing embedded multipliers. A particularly prominent example is DeepSparse \footnote{https://github.com/neuralmagic/deepsparse}, which efficiently deploys large-scale sparse models like BERT on modern Intel CPUs. DeepSparse reports up to 10× model compression with less than 1\% accuracy loss, 10× CPU inference speedup with under 2\% drop, and as much as 29× speedup with under 7.5\% accuracy degradation.

\begin{table*}[ht]
\centering
\caption{End-to-end speedup of LLaMA-V2-7B under different sparsity levels with the DeepSparse inference engine.}
\label{tab:speed}
\begin{tabular}{lcccccc}
\toprule
\textbf{Sparsity} & \textbf{Dense}  & \textbf{40\%} & \textbf{50\%} & \textbf{60\%} & \textbf{70\%} & \textbf{80\%} \\
\midrule
Latency(ms )& 206.36  & 179.22 & 113.69 & 95.18 & 82.29 & 63.52  \\
Throughput & 4.84     & 5.58 & 8.79 & 10.50 & 12.15 & 15.74  \\
Speedup & 1.0$\times$  &  1.2$\times$ & 1.8$\times$ & 2.2$\times$ & 2.5$\times$ & 3.3$\times$ \\
\bottomrule
\end{tabular}
\end{table*}

Motivated by these developments, we evaluate actual inference speedup using the DeepSparse engine~\citep{deepsparse}. Specifically, we measure end-to-end decoding latency for the LLaMA-V2-7B model running on an Intel Xeon Platinum 8360Y CPU with 36 cores. Results show that sparse LLMs fine-tuned with SEFT achieve substantial speedups over their dense counterparts, reaching up to 2.5× improvement at 70\% sparsity. Furthermore, the benefit becomes even more pronounced at higher sparsity levels, achieving approximately 4× speedup at 90\% sparsity, highlighting the promise of extreme sparsity for efficient inference. These findings underscore the importance of maintaining sparsity throughout the fine-tuning process and point toward the potential of leveraging sparsity-aware GPU operations in future deployment scenarios.

\begin{figure*}[!htb]
    \centering
    \includegraphics[width=0.9\textwidth]{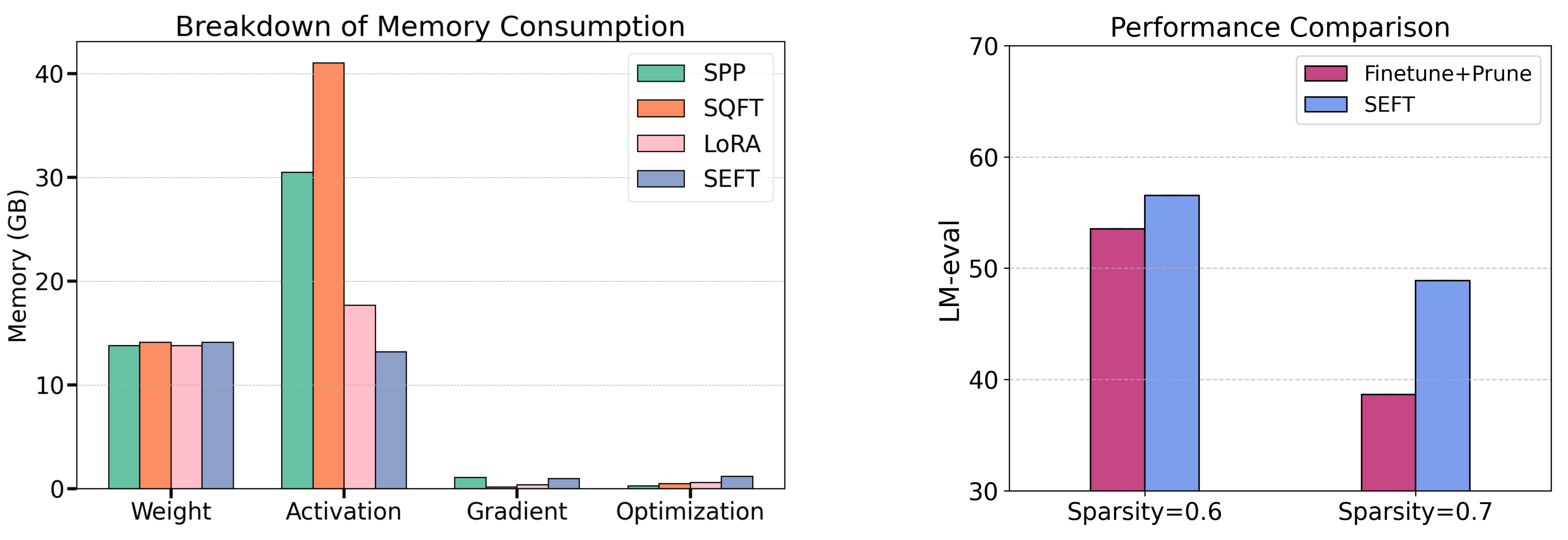}
    \caption{(a) Breakdown of memory consumption for different methods when fine-tuning the LLaMA-2 7B model. (b) LM-eval performance of LLaMA-2 7B under varying sparsity levels, comparing SEFT with baseline approaches.}
    \label{fig: app_comp}
\end{figure*}

\section{Memory Usage Comparison}
\label{sec:mem_usage}
\addcontentsline{apc}{section}{Memory Usage Comparison}

We measure end-to-end GPU memory consumption and its component-wise breakdown (parameters, activations, gradients, and optimizer states) on fine-tuning the LLaMA-2 7B model. Unless otherwise noted, we use batch size~$=1$, input sequence length~$=1024$, and \emph{disable} gradient checkpointing. For the compared methods (LoRA, SPP, and SQFT), we apply the same settings. The results are shown in Figure~\ref{fig: app_comp} (a).

SEFT exhibits substantially lower \emph{activation} memory than the baselines, whereas its \emph{gradient} and \emph{optimizer} memory are slightly higher. The latter is attributable to temporary buffers used during the sparse topology evolution step. Overall, SEFT achieves the lowest total memory footprint. Moreover, because activation memory scales roughly linearly with sequence length, SEFT’s advantage increases at longer contexts; hence we expect even larger savings as the sequence length grows.

\section{Comparison with Post-hoc Pruning}
\label{sec:pruning}
\addcontentsline{apc}{section}{Comparison with Post-hoc Pruning}

Figure~\ref{fig: app_comp} (b) reports LM-eval performance of fine-tuning LLaMA-2 7B model across sparsity levels, comparing the standard pipeline—LoRA fine-tuning on a \emph{dense} model followed by post-hoc pruning—with \emph{SEFT}, which fine-tunes an already sparse model. SEFT consistently outperforms the \textit{prune after dense fine-tuning} pipeline, with particularly large gains at higher sparsity (e.g., >10\% LM-eval points at 70\% sparsity).

These results indicate that SEFT is not only more \emph{efficient}, by maintaining a sparse base model throughout fine-tuning, but also more \emph{effective} at aligning the sparse topology with downstream tasks. This provides a clear motivation for adopting pruning \emph{and} sparse fine-tuning as an integrated strategy, rather than fine-tuning densely and pruning afterward.

\section{Effect on N:M Sparsity}
\label{sec:nm}
\addcontentsline{apc}{section}{Effect on N:M Sparsity}

\begin{wraptable}{r}{0.5\textwidth}
\vskip -0.2in
\begin{minipage}{0.5\textwidth}
\centering
\caption{Performance evaluation of LLaMA1-7B and LLaMA3-8B models on LM-eval under different N:M sparsity patterns. Higher values indicate better performance.}
\label{tab:nm}
\resizebox{0.95\linewidth}{!}{
\begin{tabular}{l|cc|cc}
\toprule[1pt]
\multirow{2}{*}{\textbf{Method}} & \multicolumn{2}{c|}{\textbf{LLaMA1-7B}} & \multicolumn{2}{c}{\textbf{LLaMA3-8B}} \\ 
\cline{2-5}
&2:4& 4:8 & 2:4 & 4:8 \\ 
\cmidrule{1-5}
LoRA* & 57.35 & 57.71 & 58.25 & 60.04 \\
SPP & 55.16 & 56.77 & 56.55 & 57.81\\
SQFT & 56.85 & 57.87 & 56.99 & 58.28\\
SEFT & \textbf{58.17} & \textbf{58.65} & \textbf{59.63} & \textbf{60.35} \\ 
 
\bottomrule[1pt]
\end{tabular}
}
\vskip -0.2in
\end{minipage}
\end{wraptable}


In addition to demonstrating the effectiveness of SEFT under unstructured pruning in previous sections, we further evaluate its applicability to structured N:M sparsity patterns, which are increasingly supported by modern hardware (e.g., NVIDIA Ampere and Hopper architectures \citep{nvidia2021ampere}). 
To enforce these structured sparsity constraints, the sparse topology evolution phase in SEFT is restricted to exploring and activating updates only within the set of currently active weights ($M_0 = 1$), preserving the required pattern.
Table~\ref{tab:nm} reports the performance of SEFT under 2:4 and 4:8 sparsity configurations on both LLaMA1-7B and LLaMA3-8B models after fine-tuning pruned models using the Wanda method. The results show that SEFT maintains or even improves performance under structured sparsity settings, outperforming other sparsity-preserving baselines. These findings extend the applicability of SEFT beyond unstructured pruning, highlighting its versatility in both algorithmic efficiency and hardware-friendly sparse training.

\section{Ablation Study}
\label{sec:ablation}
\addcontentsline{apc}{section}{Ablation Study}

We conduct a series of ablation studies to better understand the impact of key components in SEFT. Specifically, we analyze (1) the role of sparsity adaptation in maintaining the target sparsity level, and (2) the benefit of using a sensitivity-based pruning criterion over a magnitude-based one.

\begin{table}[htbp]
\centering
\caption{Performance comparison of LLaMA1-7B, LLaMA2-7B, and LLaMA3-8B models with and without sparsity adaptation  for SEFT during sparse topology evolution at a sparsity level of 60\%.}
\label{tab:maskz_zeros1}
\resizebox{0.7\linewidth}{!}{
\begin{tabular}{ll|c|c}
\toprule[1pt]
{\textbf{Method}} & & {\textbf{LM-eval}} & {\textbf{Final sparsity}} \\ 
\cmidrule{1-4}
\multirow{2}{*}{\textbf{LLaMA1-7B}} 
& w/o. adapt & 58.85 & 0.596\\
& w. adapt & 58.84 & 0.600 \\ 
\cmidrule{1-4}
\multirow{2}{*}{\textbf{LLaMA2-7B}} 
& w/o. adapt & 56.45 & 0.596 \\
& w. adapt & 56.54 & 0.600 \\ 
\cmidrule{1-4}
\multirow{2}{*}{\textbf{LLaMA3-8B}} 
& w/o. adapt & 58.60 & 0.597 \\
& w. adapt & 58.44 & 0.600  \\ 
\bottomrule[1pt]
\end{tabular}
}
\end{table}

\subsection{Effect on Sparsity Adaption} 
\label{sec:adapt}
\addcontentsline{apc}{subsection}{Effect on Sparsity Adaption}

In Section \ref{sec:adaptation}, we proposed a sparsity adaptation process to restore the model to the desired sparsity level, as sparse topology evolution in SEFT without a mask constraint may result in a denser model by reactivating previously inactive weights.
We conducted experiments to evaluate the effectiveness of sparsity adaptation on LLaMA1-7B, LLaMA2-7B, and LLaMA3-8B models pruned to 60\% sparsity using Wanda, comparing SEFT with and without sparsity adaptation. As shown in Table \ref{tab:maskz_zeros1}, the results indicate that without sparsity adaptation, the sparse LLMs become slightly denser after fine-tuning, with a sparsity level of approximately 59.6\%. In contrast, with sparsity adaptation, the models maintain their original sparsity levels while achieving comparable performance.

\begin{figure}[!htb]
    \centering
    \includegraphics[width=0.8\textwidth]{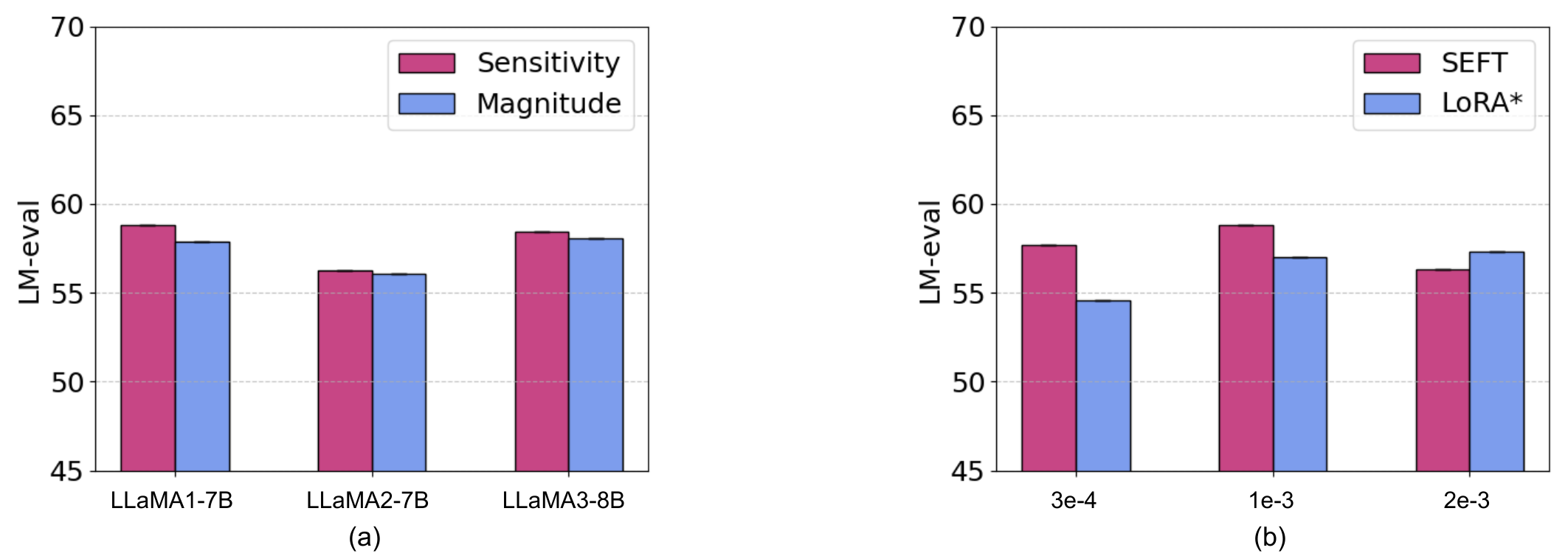}
    \caption{(a) LM-eval comparison of LLaMA1-7B, LLaMA2-7B, and LLaMA3-8B models with sensitivity-based and magnitude-based sparsity adaptation in SEFT fine-tuning at 60\% sparsity using Wanda pruning. (b) LM-eval comparison of LLaMA1-7B models at 60\% sparsity using Wanda pruning for LoRA* and SEFT fine-tuning across different learning rates.}
    \label{fig: ablation}
\end{figure}

\subsection{Importance Metrics} 
\label{sec:adapt}
\addcontentsline{apc}{subsection}{Importance Metrics}

In our work, we compare two scoring metrics: (i) \emph{magnitude-based}, \(s_i = |\theta_i|\), which is task-agnostic and inexpensive to compute; and (ii) \emph{sensitivity-based}, \(s_i = \big|\theta_i \,\nabla_{\theta_i}\mathcal{L}\big|\), which accounts for both parameter size and its contribution to the loss \citep{lee2019snip, wu2023bold, nowak2024fantastic}. 

In Figure \ref{fig: ablation} (a), we evaluate the effectiveness of sensitivity-based criterion by comparing it to the commonly used magnitude-based criterion for pruning back to the target sparsity. Magnitude-based criterion are widely used in dynamic sparse training (DST) scenarios. Our results demonstrate that sensitivity-based criterion consistently outperform magnitude-based criterion. This advantage is likely due to the sensitivity-based approach accounting not only for the magnitude of the weights but also for gradient information, which reflects their future importance during training.

\section{Sensitivity Analysis}
\label{sec:sensity}
\addcontentsline{apc}{section}{Sensitivity Analysis}

We conduct a detailed sensitivity analysis to examine how SEFT responds to different hyperparameter choices. Specifically, we study (1) the impact of sparsity levels on fine-tuning performance, (2) the effect of the number of trainable parameters, (3) the influence of drop rate and (4) evolution frequency in sparse topology evolution, and (5) the sensitivity to learning rate configurations. These analyses offer insights into SEFT's robustness and adaptability across settings, and help guide optimal hyperparameter choices for various deployment scenarios.

\subsection{Impact on Sparsity Level}
\label{sec:sparsity}
\addcontentsline{apc}{subsection}{Impact on Sparsity Level}

Table \ref{tab:sparsity_level} presents the performance evaluation of LLaMA2-7B and LLaMA3-8B models using Wanda pruning under varying sparsity levels. The results show that as the sparsity level increases, SEFT consistently outperforms LoRA* by larger margins. For instance, at a sparsity level of 70\%, SEFT achieves significant improvements in both LM-eval and MMLU scores compared to LoRA*. Notably, on LLaMA3-8B, the relative gain in MMLU reaches approximately 8.69 points (39.87 vs. 31.18).
This trend highlights that SEFT's ability to dynamically adapt the sparse topology during fine-tuning is particularly effective in high-sparsity scenarios. The primary reason for this is that post-training pruning methods like Wanda tend to degrade more significantly at higher sparsity levels, resulting in greater performance gaps. By leveraging sensitivity-based criterion, SEFT is able to better identify and optimize critical parameters, maintaining strong performance even under extreme sparsity conditions where LoRA* struggles to achieve comparable results.

\begin{figure*}[!t]
    \centering
    \includegraphics[width=\textwidth]{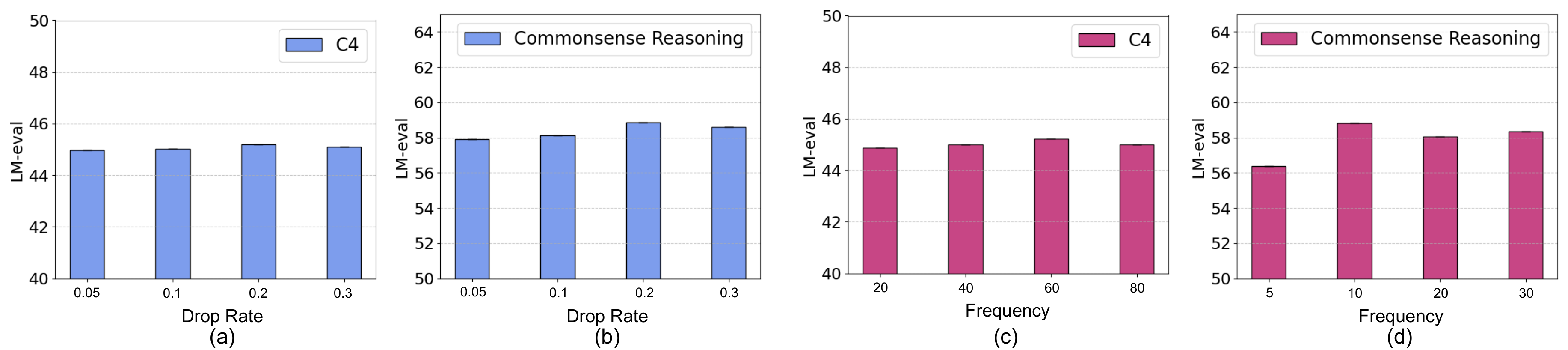}
    \vskip -0.05in
    \caption{Impact of drop rate during sparse topology evolution on LM-eval evaluation of LLaMA1-7B fine-tuning for (a) C4 and (b) commonsense reasoning datasets. Furthermore, the impact of update frequency during sparse topology evolution is evaluated on LM-eval for LLaMA1-7B fine-tuning on (c) C4 and (d) commonsense reasoning datasets.}
    \label{fig: app_ablation}
\end{figure*}

\begin{table}[htbp]
\centering
\caption{Performance comparison of LLaMA2-7B and LLaMA3-8B models on LM-eval and MMLU under different sparsity levels ($\rho$).}
\label{tab:sparsity_level}
\resizebox{0.8\linewidth}{!}{
\begin{tabular}{ll|ccc|ccc}
\toprule[1pt]
\multirow{2}{*}{\textbf{Method}} & & \multicolumn{3}{c|}{\textbf{LLaMA2-7B}} & \multicolumn{3}{c}{\textbf{LLaMA3-8B}} \\ 
\cline{3-8}
 & &$\rho$=0.5& $\rho$=0.6 & $\rho$=0.7 & $\rho$=0.5 & $\rho$=0.6 & $\rho$=0.7 \\ 
\cmidrule{1-8}
\multirow{2}{*}{\textbf{LM-eval}} 
& LoRA* &\textbf{60.12} &56.19 & 46.36 &62.90 & 55.56 & 44.82 \\
& SEFT & 59.30 & \textbf{56.54} & \textbf{48.87} & \textbf{63.04} & \textbf{58.44} & \textbf{47.44}\\ 
\cmidrule{1-8}
\multirow{2}{*}{\textbf{MMLU}} 
& LoRA* &48.21 & 42.29 &25.37 &57.40 & 49.18 & 31.18 \\
& SEFT &\textbf{48.90} & \textbf{45.95} & \textbf{33.51} &\textbf{57.51} & \textbf{50.33} & \textbf{39.87} \\ 
\bottomrule[1pt]
\end{tabular}
}
\end{table}

\subsection{Impact on Drop Rate}
\label{sec:drop}
\addcontentsline{apc}{subsection}{Impact on Drop Rate}

To investigate the effect of varying drop rates, which determine the fraction of weights pruned during each update, we conducted experiments using different initial drop rates for SEFT fine-tuning on the C4 and Commonsense Reasoning datasets. Drop rates of {0.05, 0.1, 0.2, 0.3} were evaluated, along with a cosine schedule decay for the drop/grow ratio, as proposed in \citep{evci2020rigging}.

As shown in Figure \ref{fig: app_ablation} (a) (b), the results demonstrate that a drop rate of 0.2 consistently achieves the best performance across these tasks. Lower drop rates, such as 0.05, fail to sufficiently adapt the sparse topology, limiting performance improvements, while higher drop rates, such as 0.3, can overly perturb the model, disrupting training convergence. The balanced drop rate of 0.2 was adopted as the default setting for all subsequent experiments, as it effectively balances adaptation and stability during sparse topology evolution.

\subsection{Impact on Update Frequency $k$}
\label{sec:freq}
\addcontentsline{apc}{subsection}{Impact on Update Frequency $k$}

The frequency of sparse topology evolution, which determines how often the sparse topology is updated during training, is another critical factor. We evaluated different drop/grow frequencies for fine-tuning on the C4 dataset ({10, 20, 40, 80} steps) and the Commonsense Reasoning dataset ({5, 10, 20, 30} steps).

As shown in Figure \ref{fig: app_ablation} (c) (d), the optimal frequency varies by task, reflecting differences in training dynamics and the rate at which sparse topology adaptation benefits performance. For example, on the C4 dataset, the best performance was achieved with an update frequency of 60 steps, while for the Commonsense Reasoning dataset, the optimal frequency was 10 steps. These frequencies were adopted as the default settings in our main experiments.

The results also reveal that overly frequent updates can introduce excessive instability, preventing the model from converging effectively. Conversely, infrequent updates reduce the model's ability to adapt its sparse topology, thereby limiting the benefits of sparsity. Striking the right balance in the frequency of topology evolution is crucial for maximizing SEFT's performance across diverse tasks.

\subsection{Impact on Number of Fine-tuning Parameters}
\label{sec:ranks}
\addcontentsline{apc}{subsection}{Impact on Number of Fine-tuning Parameters}

\begin{table}[htbp]
\centering
\caption{Performance comparison of LLaMA1-7B fine-tuned on the commonsense reasoning dataset at a sparsity level of 60\%. The results compare different numbers of fine-tuning parameters and report the average zero-shot accuracy across seven tasks from the commonsense reasoning benchmark.}
\label{tab:ranks}
\resizebox{0.7\linewidth}{!}{
\begin{tabular}{l|c|c|c|c}
\toprule[1pt]
{\textbf{Method}} & Rank=8 & Rank=16 & Rank=32 & Rank=64 \\ 
\cmidrule{1-5}
LoRA* & 56.81 & 57.13 & 57.32 & 57.17\\
SEFT & 57.74 & 57.61 & 58.84 & 59.69 \\ 
\bottomrule[1pt]
\end{tabular}
}
\end{table}

In this section, we analyze how the number of fine-tuning parameters impacts the performance of SEFT and LoRA* methods. The comparison is based on different numbers of fine-tuning parameters, determined by the ranks in LoRA. For SEFT, the number of fine-tuning parameters is aligned with the corresponding parameter count of LoRA ranks to ensure a fair comparison.

As shown in Table \ref{tab:ranks}, the performance of both LoRA* and SEFT improves as the number of fine-tuning parameters increases, which is expected. However, SEFT exhibits a significantly stronger upward trend compared to LoRA*. For instance, while LoRA* shows only marginal improvements at higher ranks (e.g., from 57.32 at Rank=32 to 57.17 at Rank=64), SEFT demonstrates consistent performance gains, reaching a much higher accuracy of 59.69 at Rank=64.

This trend indicates that SEFT is better equipped to utilize the additional fine-tuning capacity to adapt sparse LLMs to downstream tasks. By dynamically evolving the sparse topology during fine-tuning, SEFT effectively redistributes its parameter budget toward critical updates, enabling it to achieve more substantial improvements as the number of parameters increases. Overall, SEFT consistently outperforms LoRA* across all ranks, with the performance gap widening at higher parameter counts. These results highlight SEFT's scalability and its ability to efficiently leverage additional fine-tuning parameters for enhanced task-specific performance.

\subsection{Impact on Learning Rate}
\label{sec:lr}
\addcontentsline{apc}{subsection}{Impact on Learning Rate}

In this section, we analyze how the learning rate affects the performance of the sparse fine-tuning method SEFT and the LoRA-based fine-tuning method LoRA*. As shown in Figure \ref{fig: ablation} (b), we conduct experiments on LLaMA1-7B under a sparsity level of 0.6, using the commonsense reasoning dataset. The results report the zero-shot accuracy across seven tasks from the commonsense reasoning benchmark.
From the figure, we observe that LoRA* generally performs better with a larger learning rate, achieving the best performance at a learning rate of 2e-3. In contrast, SEFT performs best at a slightly smaller learning rate of 1e-3. Interestingly, under smaller learning rates, SEFT achieves more significant improvements compared to LoRA*.

This phenomenon may be attributed to the fundamental differences in how the two methods update sparse LLMs. SEFT directly updates the original weights of the sparse LLMs, enabling more precise control and better adaptation, especially under smaller updates at lower learning rates. This behavior is similar to dense full fine-tuning, where the original model weights are explicitly updated to reflect task-specific requirements. In contrast, LoRA-based fine-tuning approximates updates through low-rank matrices, indirectly influencing the original weights of the sparse LLMs. This indirect update mechanism allows for more flexible and larger updates, as the original weights remain intact and act as a safeguard against errors introduced by the approximation.

These results highlight the critical role of learning rate selection in optimizing the performance of sparse fine-tuning methods. To ensure a fair comparison, we performed a grid search for both fine-tuning approaches.

\section{Comparison with Original LoRA}
\label{original_lora}
\addcontentsline{apc}{section}{Comparison with Original LoRA}

In the main paper, to ensure a fair comparison, all fine-tuned LLMs were restored to the same target sparsity level after fine-tuning. In this section, we present the original results for LoRA, which do not include sparsity restoration. In other words, the fine-tuned LLMs remain dense after merging the low-rank matrices with the original model weights.

Tables \ref{tab:recovery_app} and \ref{tab:merge_app} provide a detailed comparison between SEFT and LoRA across three datasets: C4, Commonsense Reasoning. The results demonstrate that SEFT achieves comparable performance to LoRA, even when the latter uses dense connections in the LLMs. However, SEFT maintains the sparse topology at the original sparsity level, preserving the efficiency of the sparse model.

These findings underscore the effectiveness and efficiency of SEFT, as it not only matches the performance of LoRA fine-tuning with dense models but also retains the computational advantages of sparsity, making it a more practical solution for sparse fine-tuning of large language models, , as discussed in detail in Section~\ref{sec:speed}.

\begin{table}[ht]
\centering
\caption{
Performance comparison of fine-tuning methods on 70\% sparse LLaMA models.
We report average accuracy across 7 zero-shot tasks (LM-eval $\uparrow$) and perplexity (PPL $\downarrow$; lower is better).
}
\label{tab:recovery_app}
\vspace{0.1in}
\begin{tabular}{clcc}
\toprule
\textbf{Model} & \textbf{Method} & \textbf{PPL} ($\downarrow$) & \textbf{LM-eval} ($\uparrow$) \\
\midrule
\multirow{4}{*}{LLaMA V2-7B} 
& Wanda+LoRA           & 10.82 & 45.08 \\
& Wanda+SEFT           & 11.19 & 45.61 \\
\cmidrule{2-4}
& SparseGPT+LoRA       & 10.53 & 47.80 \\
& SparseGPT+SEFT       & 11.00 & 47.95 \\
\midrule
\multirow{4}{*}{LLaMA V3-8B} 
& Wanda+LoRA           & 16.12 & 44.76 \\
& Wanda+SEFT           & 16.17 & 44.55 \\
\cmidrule{2-4}
& SparseGPT+LoRA       & 17.88 & 47.19 \\
& SparseGPT+SEFT       & 15.09 & 48.89 \\
\bottomrule
\end{tabular}
\vspace{-0.15in}
\end{table}

\begin{wraptable}{r}{0.5\textwidth}
\vskip -0.2in
\begin{minipage}{0.5\textwidth}
\centering
\caption{Performance comparison of different fine-tuning methods applied to various sparse LLaMA models at a sparsity level of ($\rho=0.6$). Results include the average accuracy of zero-shot evaluation across seven tasks from the commonsense reasoning benchmark.}
\label{tab:merge_app}
\resizebox{\linewidth}{!}{
\begin{tabular}{clcc}
\toprule
\textbf{LLaMA} & \textbf{Method}  & \textbf{LM-eval} \\
\midrule
\multirow{4}{*}{V2-7B} 
& Wanda+LoRA  &   56.66 \\
& Wanda+SEFT        &    56.54\\
\cmidrule{2-3}
& SparseGPT+LoRA  &  58.24 \\
& SparseGPT+SEFT        &   58.04 \\
\midrule
\multirow{4}{*}{V3-8B} 
& Wanda+LoRA  &   59.51 \\
& Wanda+SEFT        &   58.44 \\
\cmidrule{2-3}
& SparseGPT+LoRA  &   60.66\\
& SparseGPT+SEFT        &  60.92  \\
\midrule
\multirow{4}{*}{V1-13B} 
& Wanda+LoRA  &   61.09 \\
& Wanda+SEFT        &   62.31 \\
\cmidrule{2-3}
& SparseGPT+LoRA  &   61.99\\
& SparseGPT+SEFT        &  62.01  \\
\bottomrule
\end{tabular}
}
\end{minipage}
\vskip -0.1in
\end{wraptable}

\section{Discussion} 
\addcontentsline{apc}{section}{Discussion}

Preserving sparsity during and after the fine-tuning of sparse LLMs is essential for maintaining computational efficiency, particularly in resource-constrained scenarios. Methods like LoRA, originally designed for resource-efficient fine-tuning of dense pre-trained models, fail to preserve sparsity after merging, limiting their effectiveness for sparse LLMs fine-tuning. Recent approaches, such as SPP \citep{luspp}, extend LoRA to support sparse LLMs by incorporating masking mechanisms to preserve sparsity. However, SPP enforces a fixed sparse topology across all tasks, which limits its adaptability to the specific requirements of downstream applications.

Sparse fine-tuning was first scaled to dense LLM fine-tuning in SpIEL \citep{ansell2024scaling}. Unlike LoRA-based methods, which use trainable low-rank matrices to parameterize adaptations, fine-tuning directly updates a small fraction of model weights through updates and their corresponding index vectors. In this paper, we extend this concept to sparse LLMs fine-tuning by enabling the updates to dynamically evolve and adapt in sparse LLMs. While SEFT builds upon the core concept of sparse fine-tuning, it introduces several key innovations tailored to sparse LLMs:

(1) Sparse setting. {Unlike SpIEL and SMT \citep{he2025smt}, which operates on dense models}, SEFT is designed for fine-tuning already-pruned sparse LLMs. This setting imposes strict sparsity constraints that require careful management during training.

(2) Update flexibility sparsity adaptation.  SEFT allows updates to zero-valued (previously pruned) weights, enabling recovery of important connections for downstream tasks. It also incorporates a sparsity adaptation step to ensure the model maintains its target sparsity after each update cycle—both of which are absent in SpIEL and SMT.

{(3) The main motivation of SEFT is to recover pruned connections that are relevant to specific downstream tasks. As many state-of-the-art LLM pruning methods rely heavily on calibration data and may remove connections crucial for task-specific generalization. While the pruning-and-growth mechanism in SpIEL and the parameter selection in SMT are designed to choose which parameters to update, the pruning-and-growth mechanism in SEFT is primarily used to dynamically recovery the pruned but task-relevant weights during finetuning. Additionally, SMT selects the most influential sparse sub-matrices only during the warm-up phase, whereas SEFT applies its pruning-and-growth mechanism to select and recover task-relevant parameters throughout training.}

These distinctions make SEFT a more tailored solution for fine-tuning sparse LLMs across varying sparsity levels.

\section{Limitations and Future Work} 
\label{app:limits}
\addcontentsline{apc}{section}{Limitations and Future Work}

While our work highlights the effectiveness of SEFT in fine-tuning sparse LLMs and enhancing their performance on downstream tasks, there are certain limitations that warrant further investigation.

One key limitation of SEFT lies in the need to compute full dense gradients during the sparse topology evolution phase, from which sparse delta updates are subsequently extracted. Although our implementation mitigates memory pressure by computing and applying gradients sequentially on a layer-by-layer basis and releasing them immediately afterward—thereby ensuring memory efficiency and avoiding large spikes—this design still limits computational efficiency. In particular, it does not fully leverage the potential speed and resource benefits of sparsity on GPUs, since the dense gradient matrix must first be calculated before isolating the sparse updates.

Writing an efficient CUDA kernel tailored for dynamic sparse training operations is an area of ongoing work. Such a kernel would enable direct computation of sparse updates without requiring operations on the full gradient matrix, significantly reducing memory usage and computational overhead. Once optimized, this approach could fully unlock the potential of SEFT and dynamic sparse training paradigms, making them more practical and efficient for large-scale federated and distributed learning scenarios.

\section{Impact Statements} 
\label{app:impact}
\addcontentsline{apc}{section}{Impact Statements}

This paper contributes to the development of efficient fine-tuning methods for large language models (LLMs), addressing the challenges posed by their substantial computational and performance demands and enhancing their feasibility for real-world deployment. While our contributions do not inherently lead to negative societal impacts, we encourage the community to remain mindful of potential ethical and practical implications when extending or applying our research.

\section{Experimental Settings}
\label{sec:hyper}
\addcontentsline{apc}{section}{Experimental Settings}

For training 7B models, we use a learning rate of 1e-3, except for Mistral-7B, where we use 5e-5. For 8B models, we adopt a learning rate of 3e-4. Gradient accumulation steps are set to 128, and we apply cosine learning rate decay throughout training. We use the AdamW optimizer with the default settings provided by the Transformers library, and no weight decay is applied. 
For baseline methods, we primarily build upon the official implementations of SPP\footnote{https://github.com/Lucky-Lance/SPP} and SQFT\footnote{https://github.com/IntelLabs/Hardware-Aware-Automated-Machine-Learning}.

All models are fine-tuned using post-training pruned LLMs obtained via SparseGPT and Wanda under various sparsity patterns and levels. We use fixed hyperparameters across experiments and do not perform tuning for specific sparsity configurations. We follow the dataset configurations in \citep{li2024owlore}, with details summarized in Table~\ref{tab:dataset}.

\begin{table}[ht]
\centering
\caption{Hyperparamters used of SEFT for fine-tuning on various benchmarks.}
\label{tab:dataset}
\resizebox{0.7\linewidth}{!}{
\begin{tabular}{cccc}
\toprule
{Benchmarks} & {Commonsense Reasoning}  & {MMLU} & { GSM8K}\\
\midrule
Train Samples & 170K & 99.8K & 7.4K \\
Test Samples & 22.4K & 14K & 1.3K \\
Max Length  & 512 & 512  & 512  \\
Training Epoch & 1 & 1 & 5 \\
Drop Rate & 0.2 & 0.2 & 0.2 \\
Frequency & 10 & 60 & 10 \\
\bottomrule
\end{tabular}
}
\end{table}



\end{document}